%% file: camera_ready.tex
\documentclass{article} 
\usepackage{iclr2023_conference,times}

\input{tex_files/math_commands.tex}

\usepackage[utf8]{inputenc} 
\usepackage[T1]{fontenc}    
\usepackage{booktabs}       
\usepackage{amsfonts}       
\usepackage{nicefrac}       
\usepackage{microtype}      

\usepackage[shortlabels]{enumitem}

\usepackage{comment}
\usepackage{soul}
\usepackage{amsmath,amssymb}
\usepackage{amsthm}
\usepackage{listings}
\usepackage{graphicx}
\usepackage{caption}
\usepackage{subcaption}
\usepackage{wrapfig}
\usepackage{pifont}
\usepackage{dsfont}
\usepackage[T1,hyphens]{url}
\usepackage{hyperref}
\usepackage{mathtools}
\usepackage{cleveref}
\usepackage{array}
\usepackage{float}
\usepackage{tabularx}
\usepackage{algorithm}
\usepackage{algpseudocode}
\usepackage{tablefootnote}
\usepackage{threeparttable}
\usepackage{adjustbox}
\usepackage{makecell}
\usepackage{setspace}
\usepackage{multirow}
\usepackage{colortbl}
\usepackage{graphics}  

\newtheorem{definition}{Definition}[section]
\newtheorem{prop}{Proposition}
\newtheorem*{definition*}{Definition}

\newcommand\norm[1]{\lVert#1\rVert}
\newcommand{\midsepremove}{\aboverulesep = 0mm \belowrulesep = 0mm} \newcommand{\midsepdefault}{\aboverulesep = 0.605mm \belowrulesep = 0.984mm}

\title{TANGOS: Regularizing Tabular Neural Networks through Gradient Orthogonalization and Specialization}

\newcommand*\samethanks[1][\value{footnote}]{\footnotemark[#1]}

\author{Alan Jeffares\thanks{Equal contribution} \\
University of Cambridge\\
\texttt{aj659@cam.ac.uk} \\
\And
Tennison Liu\samethanks \\
University of Cambridge\\
\texttt{tl522@cam.ac.uk} \\
\And
Jonathan Crabbé \\
University of Cambridge\\
\texttt{jc2133@cam.ac.uk} \\
\AND
Fergus Imrie \\
University of California, Los Angeles\\
\texttt{imrie@ucla.edu} \\
\And
Mihaela van der Schaar \\
University of Cambridge\\
Alan Turing Institute\\
\texttt{mv472@cam.ac.uk} \\
}

\iclrfinalcopy 
\begin{document}

\maketitle

\begin{abstract} 
Despite their success with unstructured data, deep neural networks are not yet a panacea for structured tabular data. In the tabular domain, their efficiency crucially relies on various forms of regularization to prevent overfitting and provide strong generalization performance. Existing regularization techniques include broad modelling decisions such as choice of architecture, loss functions, and optimization methods. In this work, we introduce Tabular Neural Gradient Orthogonalization and Specialization (\texttt{TANGOS}), a novel framework for regularization in the tabular setting built on latent unit attributions. The gradient attribution of an activation with respect to a given input feature suggests how the neuron \emph{attends} to that feature, and is often employed to interpret the predictions of deep networks. In \texttt{TANGOS}, we take a different approach and incorporate neuron attributions directly into training to encourage orthogonalization and specialization of \emph{latent attributions} in a fully-connected network. Our regularizer encourages neurons to focus on sparse, non-overlapping input features and results in a set of diverse and specialized latent units. In the tabular domain, we demonstrate that our approach can lead to improved out-of-sample generalization performance, outperforming other popular regularization methods. We provide insight into \textit{why} our regularizer is effective and demonstrate that \texttt{TANGOS} can be applied jointly with existing methods to achieve even greater generalization performance.

\end{abstract}

\section{Introduction}

Despite its relative under-representation in deep learning research, tabular data is ubiquitous in many salient application areas including medicine, finance, climate science, and economics. Beyond raw performance gains, deep learning provides a number of promising advantages over non-neural methods including multi-modal learning, meta-learning, and certain interpretability methods, which we expand upon in depth in \Cref{app:motivation}. Additionally, it is a domain in which general-purpose regularizers are of particular importance. Unlike areas such as computer vision or natural language processing, architectures for tabular data generally do not exploit the inherent structure in the input features (i.e. locality in images and sequential text, respectively) and lack the resulting inductive biases in their design. Consequentially, improvement over non-neural ensemble methods has been less pervasive. Regularization methods that implicitly or explicitly encode inductive biases thus play a more significant role. Furthermore, adapting successful strategies from the ensemble literature to neural networks may provide a path to success in the tabular domain (e.g. \citealp{wen2020batchensemble}). Recent work in \cite{kadra2021well} has demonstrated that suitable regularization is essential to outperforming such methods and, furthermore, a balanced \textit{cocktail} of regularizers results in neural network superiority.

Regularization methods employed in practice can be categorized into those that prevent overfitting through data augmentation \citep{krizhevsky2012imagenet, zhang2017mixup}, network architecture choices \citep{hinton2012improving, ioffe2015batch}, and penalty terms that explicitly influence parameter learning \citep{hoerl1970ridge, tibshirani1996regression, jin2020does}, to name just a few. While all such methods are unified in attempting to improve out-of-sample generalization, this is often achieved in vastly different ways. For example, $L1$ and $L2$ penalties favor sparsity and shrinkage, respectively, on model weights, thus choosing more parsimonious solutions. Data perturbation techniques, on the other hand, encourage smoothness in the system assuming that small perturbations in the input should not result in large changes in the output. Which method works best for a given task is generally not known \emph{a priori} and considering different classes of regularizer is recommended in practice. Furthermore, combining multiple forms of regularization simultaneously is often effective, especially in lower data regimes (see e.g. \citealp{brigato2021close} and \citealp{hu2017frankenstein}). 

\begin{figure}[t!]
  \includegraphics[width=1\linewidth]{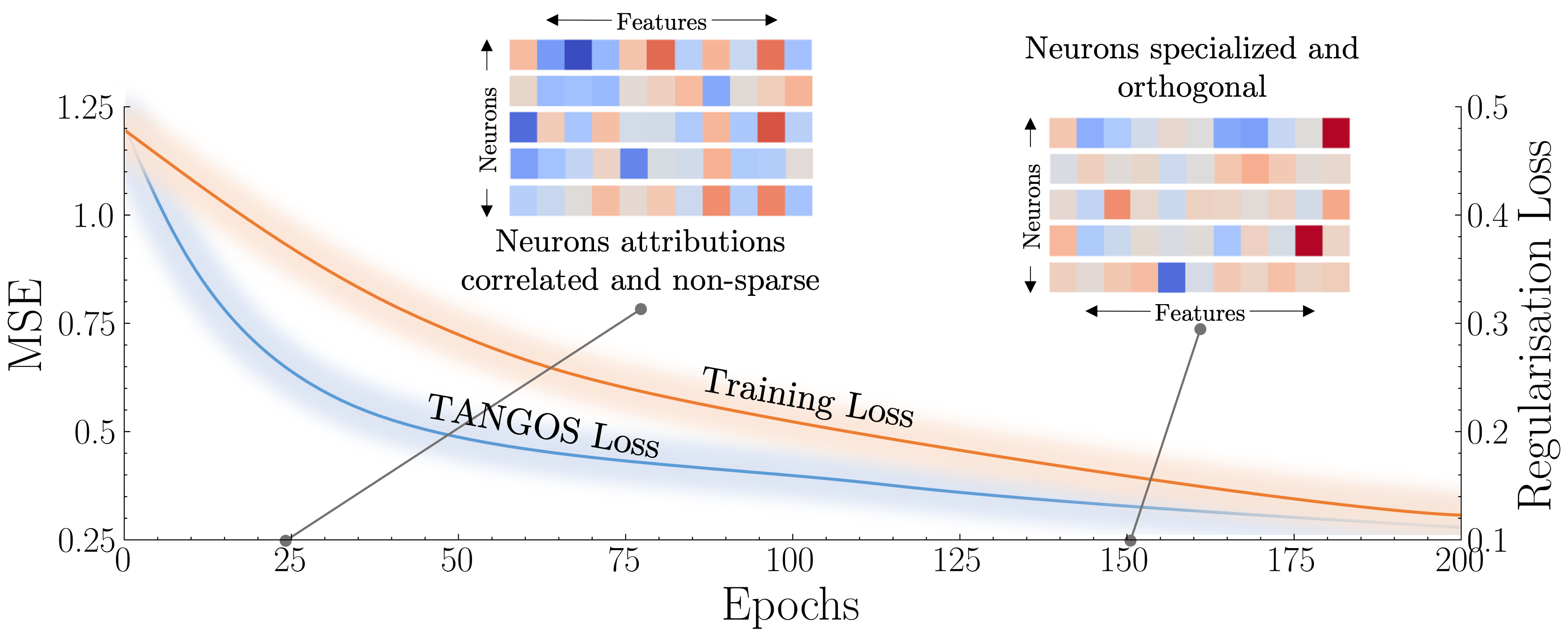}
  \vspace{-1em}
  \caption{\small \textbf{TANGOS encourages specialization and orthogonalization.} \texttt{TANGOS} penalizes neuron attributions during training. Here, \raisebox{3pt}{\fcolorbox{black}{black!25!red!75!}{\rule{0pt}{1pt}\rule{1pt}{0pt}}} indicates strong positive attribution and \raisebox{3pt}{\fcolorbox{black}{black!25!blue!75!}{\rule{0pt}{1pt}\rule{1pt}{0pt}}} indicates strong negative attribution, while interpolating colors reflect weaker attributions. Neurons are regularized to be \emph{specialized} (attend to sparser features) and \emph{orthogonal} (attend to non-overlapping features).}
  \label{fig:p1_fig}
  \vspace{-1em}
\end{figure}

Neuroscience research has suggested that neurons are both \emph{selective} \citep{johnston1986selective} and have \emph{limited capacity} \citep{cowan2005capacity} in reacting to specific physiological stimuli. Specifically, neurons selectively choose to focus on a few chunks of information in the input stimulus. In deep learning, a similar concept, commonly described as a \emph{receptive field}, is employed in convolutional layers \citep{luo2016understanding}. Here, each convolutional unit has multiple filters, and each filter is only sensitive to specialized features in a local region. The output of the filter will activate more strongly if the feature is present. This stands in contrast to fully-connected networks, where the all-to-all relationships between neurons mean each unit depends on the entire input to the network. We leverage this insight to propose a regularization method that can encourage artificial neurons to be more specialized and orthogonal to each other.

\textbf{Contributions.} \textbf{(1) Novel regularization method for deep tabular models.} In this work, we propose \texttt{TANGOS}, a novel method based on regularizing neuron attributions. A visual depiction is given in Figure \ref{fig:p1_fig}. 
Specifically, each neuron is more \emph{specialized}, attending to sparse input features while its attributions are more \emph{orthogonal} to those of other neurons. In effect, different neurons pay attention to non-overlapping subsets of input features resulting in better generalization performance. We demonstrate that this novel regularization method results in excellent generalization performance on tabular data when compared to other popular regularizers.
\textbf{(2) Distinct regularization objective.} We explore how \texttt{TANGOS} results in distinct emergent characteristics in the model weights. We further show that its improved performance is linked to increased diversity among weak learners in an ensemble of latent units, which is generally in contrast to existing regularizers.
\textbf{(3) Combination with other regularizers.} Based upon these insights, we demonstrate that deploying \texttt{TANGOS} \emph{in tandem} with other regularizers can further improve generalization of neural networks in the tabular setting beyond that of any individual regularizer.

\section{Related Work}
\textbf{Gradient Attribution Regularization.} 
A number of methods exist which incorporate a regularisation term to penalize the network gradients in some way.
Penalizing gradient attributions is a natural approach for achieving various desirable properties in a neural network. Such methods have been in use at least since \cite{drucker1992improving}, where the authors improve robustness by encouraging invariance to small perturbations in the input space. More recently, gradient attribution regularization has been successfully applied across a broad range of application areas. Some notable examples include encouraging the learning of robust features in auto-encoders \citep{rifai2011contractive}, improving stability in the training of generative adversarial networks \citep{gulrajani2017improved}, and providing robustness to adversarial perturbations \citep{moosavi2019robustness}. While many works have applied a shrinkage penalty (L2) to input gradients, \citet{ross2017neural} explore the effects of encouraging sparsity by considering an L1 penalty term. Gradient penalties may also be leveraged to compel a network to \textit{attend} to particular human-annotated input features \citep{ross2017right}. A related line of work considers the use of gradient aggregation methods such as Integrated Gradients \citep{sundararajan2017axiomatic} and, typically, penalizes their deviation from a given target value (see e.g. \citet{liu2019incorporating} and \citet{chen2019robust}). In contrast to these works, we do not require manually annotated regions upon which we constrain the network to attend. Similarly, \citet{erion2021improving} provide methods for encoding domain knowledge such as smoothness between adjacent pixels in an image. We note that while these works have investigated penalizing a predictive model's output attributions, we are the first to regularize attributions on latent neuron activations. We provide an extended discussion of related works on neural network regularization more generally in \Cref{app:related-works}.

\section{TANGOS} 

\subsection{Problem Formulation}
We operate in the standard supervised learning setting, with $d_X$-dimensional input variables $X \in \mathcal{X} \subseteq \mathbb{R}^{d_X}$ and target output variable $Y \in \mathcal{Y}\subseteq \mathbb{R}$. Let $P_{XY}$ denote the joint distribution between input and target variables. The goal of the supervised learning algorithm is to find a predictive model, $f_\theta:\mathcal{X}\rightarrow\mathcal{Y}$ with learnable parameters $\theta \in \Theta$. The predictive model belongs to a hypothesis space $f_\theta \in \mathcal{H}$ that can map from the input space to the output space. 

The predictive function is usually learned by optimizing a  loss function $\mathcal{L}: \Theta \rightarrow \mathbb{R}$ using \emph{empirical risk minimization} (ERM). The empirical risk cannot be directly minimized since the data distribution $P_{XY}$ is not known. Instead, we use a finite number of iid samples $(x, y)\sim P_{XY}$, which we refer to as the training data $\mathcal{D}=\{(x_i, y_i)\}_{i=1}^N$. 

Once the predictive model is trained on $\mathcal{D}$, it should ideally predict well on out-of-sample data generated from the same distribution. However, overfitting can occur if the hypothesis space $\mathcal{H}$ is too complex and the sampling of training data does not fully represent the underlying distribution $P_{XY}$. Regularization is an approach that reduces the complexity of the hypothesis space so that more generalized functions are learned to explain the data. This leads to the following ERM:
\begin{equation}
	\theta^* = \argmin_{\theta \in \Theta} \frac{1}{|\mathcal{D}|}\sum_{(x, y) \in \mathcal{D}} \mathcal{L}(f_\theta(x), y) + \mathcal{R}(\theta, x, y),
\end{equation}
that includes an additional regularization term $\mathcal{R}$ which, generally, is a function of input $x$, the label $y$, the model parameters $\theta$, and reflects prior assumptions about the model. For example, $L1$ regularization reflects the belief that sparse solutions in parameter space are more desirable.

\subsection{Neuron Attributions}
Formally, attribution methods aim to uncover the importance of each input feature of a given sample to the prediction of the neural network. Recent works have demonstrated that feature attribution methods can be incorporated into the training process \citep{lundberg2017unified, erion2021improving}. These \emph{attribution priors} optimize attributions to have desirable characteristics, including interpretability as well as smoothness and sparsity in predictions. However, these methods have exclusively investigated \emph{output} attributions, i.e., contributions of input features to the output of a model. To the best of our knowledge, we are the first work to investigate regularization of \emph{latent attributions}.

\begin{wrapfigure}{r!}{0.38\textwidth}
    \begin{center}
        \includegraphics[width=0.38\textwidth]{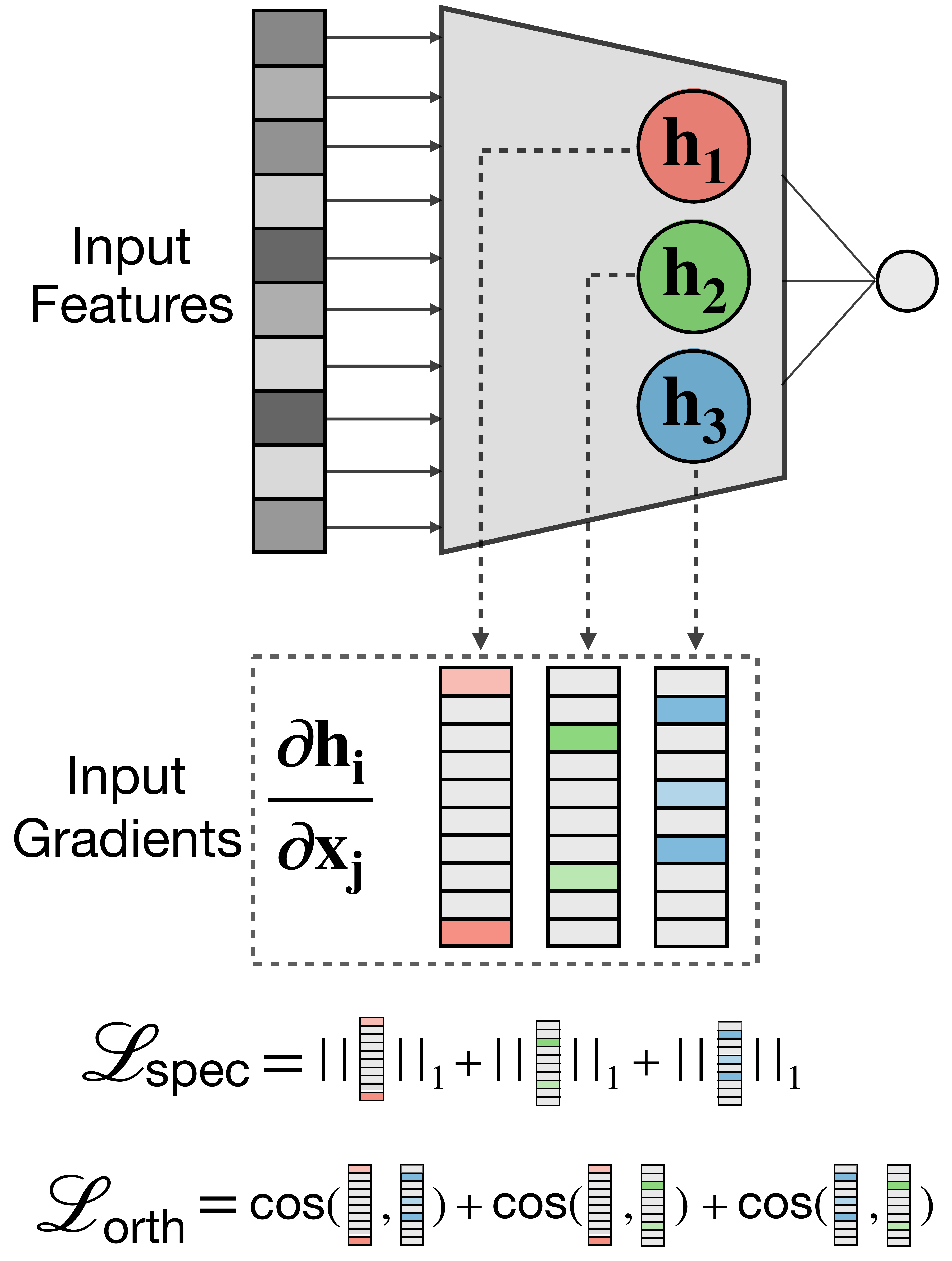}
    \end{center}
    \caption{\small \textbf{Method illustration.} \texttt{TANGOS} regularizes the gradients with respect to each of the latent units.}
    \label{fig:blah}
\vspace{-2em}
\end{wrapfigure}

We rewrite our predictive function $f$ using function composition $f = l \circ g$. Here $g: \mathcal{X}\rightarrow \mathcal{H}$ maps the input to a representation $h=g(x) \in \mathcal{H}$, where $\mathcal{H}\subseteq \mathbb{R}^{d_H}$ is a $d_H$-dimensional latent space. Additionally, $l: \mathcal{H}\rightarrow\mathcal{Y}$ maps the latent representation to a label space $y=l(h) \in \mathcal{Y}$. We let $h_i = g_i(x)$, for, $i \in [d_H]$ denote the $i^{th}$ neuron in the hidden layer of interest. Additionally, we use $a^i_j(x) \in \mathbb{R}$ to denote the attribution of the $i^{th}$ neuron w.r.t. the feature $x_j$. With this notation, upper indices correspond to latent units and lower indices to features. In some cases, it will be convenient to stack all the feature attributions together in the attribution vector $a^i(x) = [a^i_j(x)]_{j=1}^{d_X} \in \mathbb{R}^{d_X}$.

Attribution methods work by using gradient signals to evaluate the contributions of the input features. In the most simplistic setting:
\begin{equation}
    a^i_j(x) \equiv \frac{\partial h_i(x)}{\partial x_j}.
\end{equation}
This admits a simple interpretation through a first-order Taylor expansion: if the input feature $x_j$ were to increase by some small number $\epsilon \in \mathbb{R}^+$, the neuron activation would change by $\epsilon \cdot a^i_j(x) + \mathcal{O}(\epsilon^2)$. The larger the absolute value of the gradient, the stronger the effect of a change in the input feature. We emphasize that our method is \emph{agnostic} to the gradient attribution method, as different methods may be more appropriate for different tasks. For a comprehensive review of different methods, assumptions, and trade-offs, see \cite{ancona2017towards}. For completeness, we also note another category of attribution methods is built around \emph{perturbations}: this class of methods evaluates contributions of individual features through repeated perturbations. Generally speaking, they are more computationally inefficient due to the multiple forward passes through the neural network and are difficult to include directly in the training objective.

\subsection{Rewarding Orthogonalization and Specialization}
The main contribution of this work is proposing regularization on neuron attributions. In the most general sense, any function of any neuron attribution method could be used as a regularization term, thus encoding prior knowledge about the properties a model should have. 

Specifically, the regularization term is a function of the network parameters $\theta$ and $x$, i.e., $\mathcal{R}(\theta, x)$, and encourages prior assumptions on desired behavior of the learned function. Biological sensory neurons are highly specialized. For example, certain visual neurons respond to a specific set of visual features including edges and orientations within a single receptive field. They are thus highly \emph{selective} with \emph{limited capacity} to react to specific physiological stimuli \citep{johnston1986selective, cowan2005capacity}. Similarly, we hypothesize that neurons that are more specialized and pay attention to sparser signals should exhibit better generalization performance. We propose the following desiderata and corresponding regularization terms:

\begin{itemize}[leftmargin=5.0mm]
    \item \textbf{Specialization.} The contribution of input features to the activation of a particular neuron should be sparse, i.e., $||a^i(x)||$ is small for all $i \in [d_H]$ and $x \in \mathcal{X}$. Intuitively, in higher-dimensional settings, a few features should account for a large percentage of total attributions while others are near zero, resulting in more \emph{specialized} neurons. We write this as a regularization term for mini-batch training:
    \begin{equation*}
        \mathcal{L}_{\mathrm{spec}}(x) = \frac{1}{B}\sum_{b=1}^B\frac{1}{d_H}\sum_{i=1}^{d_H} \norm{a^i(x_b)}_1, \\
    \end{equation*}
    where $b \in [B]$ is the batch index of $x_b \in \mathcal{X}$ and $\norm{\cdot}_1$ denotes the $l_1$ norm.
    
    \item \textbf{Orthogonalization.} Different neurons should attend to non-overlapping subsets of input features given a particular input sample. To encourage this, we penalize the correlation between neuron attributions $\rho[a^i(x), a^j(x)]$ for all $i \neq j$ and $x \in \mathcal{X}$. In other words, for each particular input, we want to discipline the latent units to attend to different aspects of the input.  Then, expressing this as a regularization term for mini-batch training, we obtain:
    \begin{equation*}
        \mathcal{L}_{\mathrm{orth}}(x) = \frac{1}{B} \sum_{b=1}^B\frac{1}{C}\sum_{i=2}^{d_H}\sum_{j=1}^{i-1}\rho \left[a^i(x_b), a^j(x_b)\right].
    \end{equation*}
    Here, $C$ is the number of pairwise correlations, $C=\frac{d_H\cdot(d_H-1)}{2}$, and $\rho[a^i(x_b), a^j(x_b)] \in [0, 1]$ is calculated using the cosine similarity $\frac{|a^{i \intercal}(x_b) \ a^j(x_b)|}{||a^i(x_b)||_2||a^j(x_b)||_2}$ where $\norm{\cdot}_2$ denotes the $l_2$ norm.
\end{itemize}

These terms can be combined into a single regularization term and incorporated into the training objective. The resulting \texttt{TANGOS} regularizer can be expressed as:
\begin{equation*}
    \mathcal{R}_{\texttt{TANGOS}}(x) = \lambda_1 \mathcal{L}_{\mathrm{spec}}(x) + \lambda_2 \mathcal{L}_{\mathrm{orth}}(x),
\end{equation*}
where $\lambda_1$, $\lambda_2$ $\in \mathbb{R}$ act as weighting terms. As this expression is computed using gradient signals, it can be efficiently implemented and minimized in any auto-grad framework.

\section{\emph{How} and \emph{Why} Does TANGOS Work?}
To the best of our knowledge, \texttt{TANGOS} is the only work to explicitly regularize latent neuron attributions. A natural question to ask is (1) \textit{How is} \texttt{TANGOS} \textit{different from other regularization?} While intuitively it makes sense to enforce \emph{specialization} of each unit and \emph{orthogonalization} between units, we empirically investigate if other regularizers can achieve similar effects, revealing that our method regularizes a unique objective. Having established that the \texttt{TANGOS} objective is unique, the next question is (2) \textit{Why does it work?} To investigate this question, we frame the set of neurons as an ensemble, and demonstrate that our regularization improves diversity among \emph{weak learners}, resulting in improved out-of-sample generalization. 

\vspace{-0.25em}
\subsection{TANGOS Regularizes a Unique Objective}
\begin{figure}[t!]
    \vspace{-2.5em}
    \centering
    \begin{subfigure}[b]{0.85\textwidth}
       \includegraphics[width=1\linewidth]{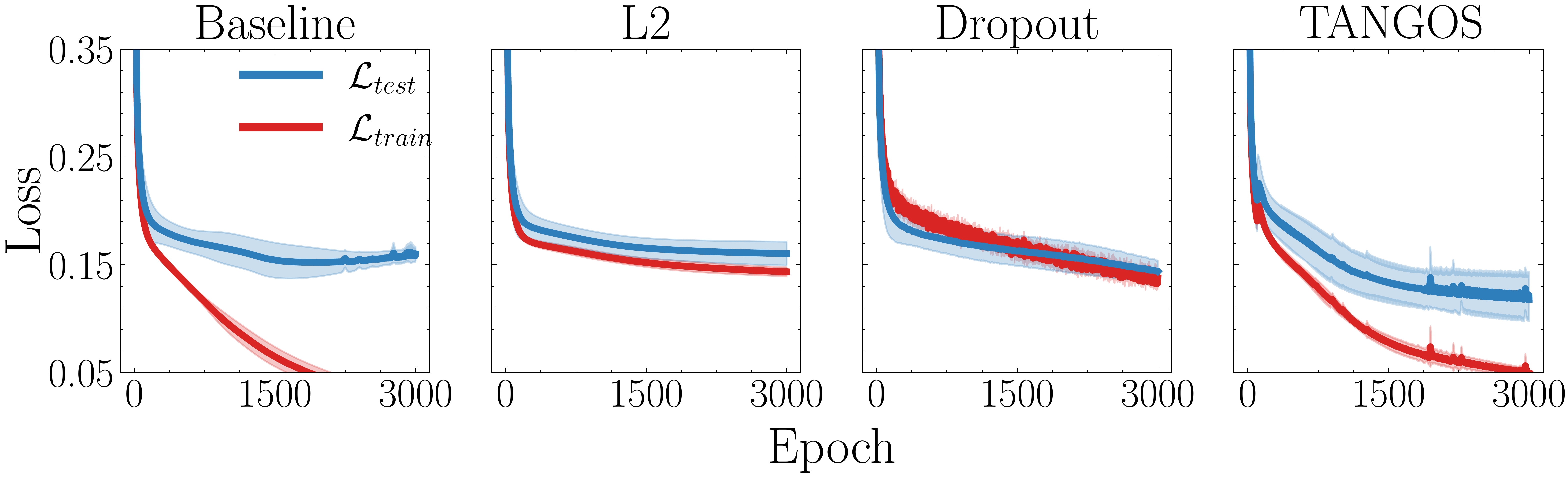}
       \label{fig:train_test_loss} 
    \end{subfigure}
    \vspace{-1em}
    \begin{subfigure}[b]{0.85\textwidth}
       \includegraphics[width=1\linewidth]{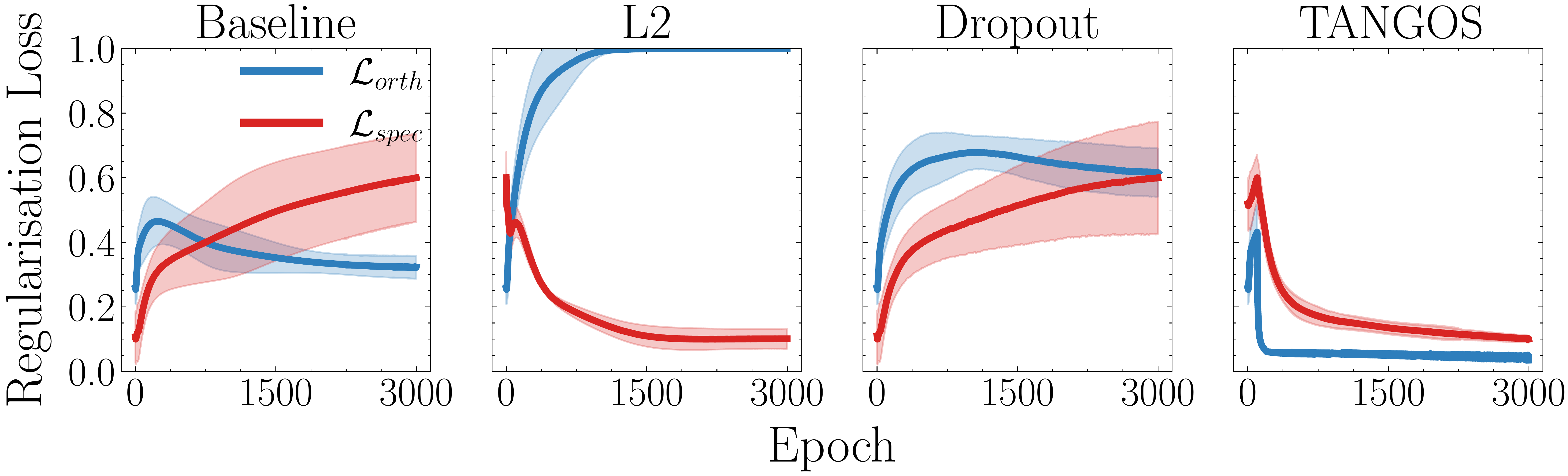}
       \label{fig:legos_reg_loss}
    \end{subfigure}
    \vspace{-1em}
    \caption{\small \textbf{Comparison of regularization objectives.} \textbf{(Top)} Generalization performance of key regularization techniques, \textbf{(Bottom)} corresponding neuron attributions evaluated on the test set. $L2$ and $DO$ can reduce overfitting, but neuron attributions are in fact becoming more correlated. \texttt{TANGOS} achieves the best generalization performance by penalizing a different objective.}
    \label{fig:comparison_regularisation}
    \vspace{-1em}
\end{figure}

\vspace{-0.25em}
\texttt{TANGOS} encourages generalization by explicitly decorrelating and sparsifying the attributions of latent units. A reasonable question to ask is if this is unique, or if other regularizers might achieve the same objective implicitly. Two alternative regularizers that one might consider are $L2$ weight regularization and $Dropout$. Like \texttt{TANGOS}, weight regularization methods implicitly and partially penalize the gradients by shrinking the weights in the neural network. Additionally, $Dropout$ trains an ensemble of learners by forcing each neuron to be more independent. In Figure \ref{fig:comparison_regularisation}, we provide these results on the UCI temperature forecast dataset \citep{cho2020comparative}, in which data from 25 weather stations in South Korea is used to predict next-day peak temperature. We train a fully connected neural network for each regularization method. Specifically, we plot $\mathcal{L}_{spec}$ and $\mathcal{L}_{orth}$ for neurons in the penultimate layers and the corresponding generalization performance. We supply an extended selection of these results on additional datasets and regularizers in Appendix~\ref{appdx:extended_results}.

First, we observe that \texttt{TANGOS} significantly decreases correlation between different neuron attributions while other regularization terms, in fact, increase them. For $L2$ weight regularization, this suggests that as the neural network weights are made smaller, the neurons increasingly attend to the same input features. A similar effect is observed for $Dropout$ - which has a logical explanation. Indeed, $Dropout$ creates redundancy by forcing each latent unit to be independent of others. 
Naturally, this encourages individual neurons to attend to overlapping features. In contrast, \texttt{TANGOS} aims to achieve specialization, such that neurons pay attention to sparse, non-overlapping features.

Additionally, we note that no alternative regularizers achieve greater attribution sparsity. This does not come as a surprise for $Dropout$, where the aim to induce redundancy in each neuron will naturally encourage individual neurons to attend to more features. While $L2$ does achieve a similar level of sparsity, this is paired with a high $\mathcal{L}_{orth}$ term indicating that, although the latent units do attend to sparse features, they appear to collapse to a solution in which they all attend to the same weighted subset of the input features. This, as we will discover in \S4.2, is unlikely to be optimal for out-of-sample generalization.

Therefore, we conclude that the pairing of the specialization and orthogonality objectives in \texttt{TANGOS} regularizes a unique objective.

\vspace{-0.25em}
\subsection{TANGOS Generalizes by Increasing Diversity Among Latent Units} \label{sec:diversity} 
\vspace{-0.25em}
Having established how \texttt{TANGOS} differs from existing regularization objectives, we now turn to answer \emph{why} it works. In this section, we provide an alternative perspective on the effect of \texttt{TANGOS} regularization in the context of ensemble learning. A predictive model $f(x)$ may be considered as an ensemble model if it can be written in the form $f(x) = \sum_{{T}_k \in \mathcal{T}}\alpha_kT_k(x)$, where $\mathcal{T}$ represents a set of basis functions sometimes referred to as \textit{weak learners} and the $\alpha_k$'s represent their respective scalar weights. It is therefore clear that each output of a typical neural network may be considered an ensemble predictor with every latent unit in its penultimate layer acting as a weak learner in their contribution to the model's output. More formally, in this setting $T_k(x)$ is the activation of latent unit $k$ with respect to an input $x$ and $\alpha_k$ is the subsequent connection to the output activation. With this in mind, we present the following definition.

\begin{definition}
    Consider an ensemble regressor $f(x) = \sum_{{T}_k \in \mathcal{T}}\alpha_kT_k(x)$ trained on $\mathcal{D}=\{(x_i, y_i)\}_{i=1}^N$ where each $(x, y)$ is drawn randomly from $P_{XY}$. Additionally, the weights are constrained such that $\sum_k\alpha_k = 1$. Then, for a given input-label pair $(x, y)$, we define:
    \begin{enumerate}[(a)]
        \item The overall ensemble error as: $\mathrm{Err} = (f(x) - y)^2$.
        \item The weighted errors of the weak learners as: $\overline{\mathrm{Err}} = \sum_k \alpha_k(T_k(x) - y)^2$.
        \item The ensemble diversity as: $\mathrm{Div} = \sum_k \alpha_k(T_k(x) - f(x))^2$.
    \end{enumerate}
\end{definition}

Intuitively, $\overline{\text{Err}}$ provides a measure of the strength of the ensemble members while $\text{Div}$ measures the diversity of their outputs. To understand the relationship between these two terms and the overall ensemble performance, we consider Proposition \ref{prop:ensemble}.

\begin{prop}[\citealt{krogh1994neural}]
    The overall ensemble error for an input-label pair $(x,y)$ can be decomposed into the weighted errors of the weak learners and the ensemble diversity such that:
    \begin{equation}
        \mathrm{Err} = \overline{\mathrm{Err}} - \mathrm{Div}.
    \end{equation}
    \label{prop:ensemble}
\end{prop}
\vspace{-2.5em}
\begin{figure}
  \centering
  \includegraphics[width=0.85\linewidth]{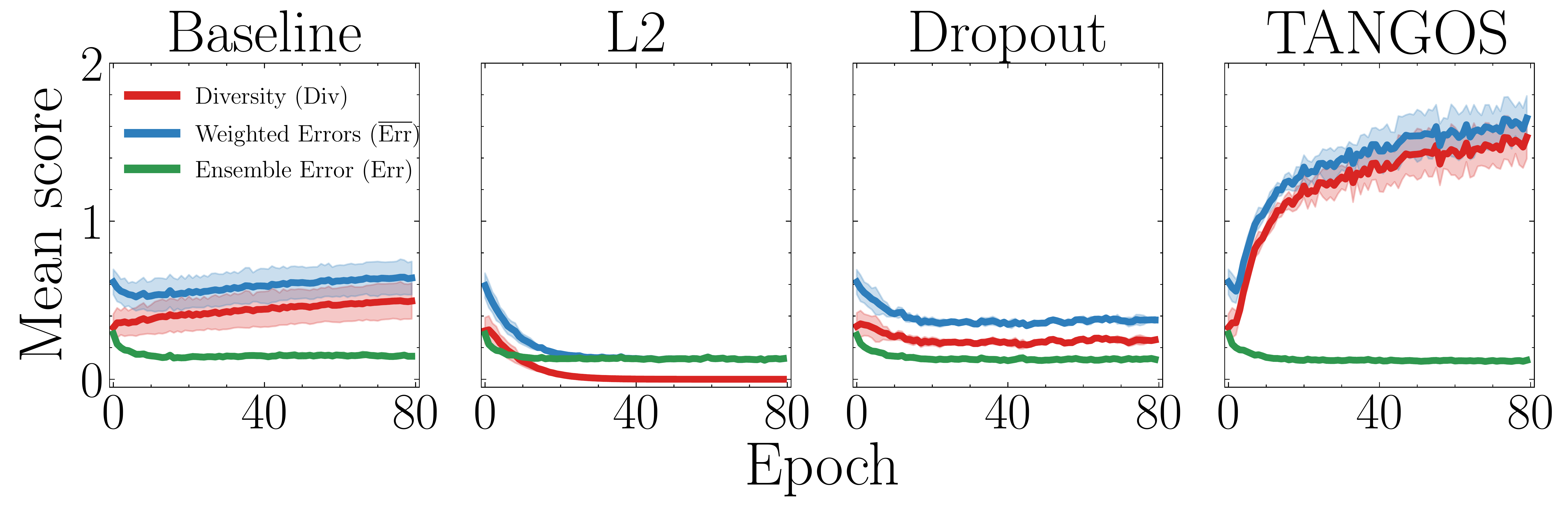}
  \vspace{-1em}
  \caption{\small \textbf{Neuron Diversity.} Overall ensemble error and decomposition in terms of diversity and average error of the weak learners. Note while all methods achieve low overall error, \texttt{TANGOS} is the only method that does so by increasing the diversity among the latent units.}
  \label{fig:diversity}
  \vspace{-1em}
\end{figure}
This decomposition provides a fundamental insight into the success of ensemble methods: an ensemble's overall error is reduced by decreasing the average error of the individual weak learners and increasing the diversity of their outputs. Successful ensemble methods explicitly increase ensemble diversity when training weak learners by, for example, sub-sampling input features (random forest, \citealp{breiman2001random}), sub-sampling from the training data (bagging, \citealp{breiman1996bagging}) or error-weighted input importance (boosting, \citealp{buhlmann2012bagging}). 

Returning to the specific case of neural networks, it is clear that \texttt{TANGOS} provides a similar mechanism of increasing diversity among the latent units that act as weak learners in the penultimate layer. By forcing the latent units to attend to sparse, uncorrelated selections of features, the learned ensemble is encouraged to produce diverse learners whilst maintaining coverage of the entire input space in aggregate. In Figure \ref{fig:diversity}, we demonstrate this phenomenon in practice by returning to the UCI temperature forecast regression task. We provide extended results in Appendix \ref{appdx:extended_results}. We train a fully connected neural network with two hidden layers with the output layer weights constrained such that they sum to 1. We observe that regularizing with \texttt{TANGOS} increases diversity of the latent activations resulting in improved out-of-sample generalization. This is in contrast to other typical regularization approaches which also improve model performance, but exclusively by attempting to reduce the error of the individual ensemble members. 
This provides additional motivation for applying \texttt{TANGOS} in the tabular domain, an area where traditional ensemble methods have performed particularly well.

\begin{table}[h!]
\vspace{-0.5em}
\tiny
\centering
\caption{\small \textbf{Stand-Alone Regularization.} Comparison of regularizers on regression and classification in terms of test MSE and NLL. All models are trained on real-world datasets using 5-fold cross-validation and final evaluation reported on a held-out test set. \textbf{Bold} indicates the best performance. The average rank of each method across both regression and classification is included in the final row of the respective tables.}
\label{tab:stand_alone_generalisation}
\midsepremove
\resizebox{0.9\textwidth}{!}{
\begin{tabular}{|c|c|c|c|c|c|c|c|c|}
\toprule
Dataset & Baseline & L1 & L2 & DO & BN & IN & MU & TANGOS \\ \midrule 
\multicolumn{9}{c}{\cellcolor[HTML]{EFEFEF}\rule{0pt}{2ex} Regression (Mean Squared Error)}                                                   \\ \midrule
FB      &  0.037  &  0.081  &  \textbf{0.029}  &  0.060  &  0.699  &  0.043  &  0.147  &  0.032  \\
BH      &  0.192  &  0.197  &  0.183  &  0.209  &  0.190  &  0.215  &  0.286  &  \textbf{0.166}  \\
WE      &  0.118  &  0.096  &  0.099  &  0.097  &  \textbf{0.090}  &  0.101  &  0.146  &  0.093  \\
BC     &  0.323  &  0.263  &  0.277  &  0.282  &  0.294  &  0.308  &  0.323  &  \textbf{0.244}  \\
WQ      &  0.673  &  0.641  &  0.644  &  0.658  &  0.639  &  0.669  &  0.713  &  \textbf{0.637}  \\
SC      &  0.422  &  0.408  &  0.411  &  0.423  &  0.410  &  0.434  &  0.547  &  \textbf{0.387}  \\
FF      &  1.274  &  1.280  &  1.274  &  1.266  &  1.330  &  \textbf{1.201}  &  1.289  &  1.276  \\
PR      &  0.624  &  0.611  &  0.580  &  0.592  &  0.647  &  0.591  &  0.745  &  \textbf{0.573}  \\
ST      &  0.419  &  0.416  &  0.418  &  0.387  &  0.461  &  0.539  &  \textbf{0.380}  &  0.382  \\
AB      &  0.345  &  0.319  &  0.332  &  \textbf{0.312}  &  0.348  &  0.355  &  0.366  &  0.325  \\
\hline \hline
Avg Rank      &  5.4  &  3.8  &  3.4  &  4.0  &  5.0   &  5.5  &  7.1  &   \textbf{1.9} \\

\midrule
\multicolumn{9}{c}{\cellcolor[HTML]{EFEFEF}\rule{0pt}{2.5ex}  Classification (Mean Negative Log-likelihood)}                                           \\ \midrule
HE     &  0.490  &  0.472  &  0.431  &  0.428  &  0.459  &  0.435  &  \textbf{0.416}  &  0.426  \\
BR     &  0.074  &  0.070  &  0.070  &  0.078  &  0.080  &  0.071  &  0.095  &  \textbf{0.069}  \\
CE     &  0.519  &  \textbf{0.395}  &  0.407  &  0.436  &  0.604  &  0.457  &  0.472  &  0.408  \\
CR     &  0.464  &  0.405  &  0.402  &  0.456  &  0.460  &  0.481  &  0.448  &  \textbf{0.369}  \\
HC     &  0.320  &  0.222  &  0.226  &  0.237  &  0.257  &  0.312  &  0.248  &  \textbf{0.215}  \\
AU     &  0.448  &  0.442  &  0.385  &  0.405  &  0.549  &  0.479  &  0.478  &  \textbf{0.379}  \\
TU     &  1.649  &  1.633  &  1.613  &  1.621  &  \textbf{1.484}  &  1.646  &  1.657  &  1.495  \\
EN     &  1.040  &  1.040  &  1.042  &  1.058  &  1.098  &  1.072  &  1.065  &  \textbf{0.974}  \\
TH     &  0.700  &  0.506  &  \textbf{0.500}  &  0.714  &  0.785  &  0.638  &  0.618  &  0.513  \\
SO     &  0.606  &  \textbf{0.238}  &  0.382  &  0.567  &  0.484  &  0.540  &  0.412  &  0.371  \\
\hline \hline
Avg Rank      &  6.4  &  3.0  &  2.7  &  4.8  &  6.3   &  6.0  &  5.2  &   \textbf{1.7} \\
\bottomrule
\end{tabular}%
}
\midsepdefault
\vspace{-1em}
\end{table}

\section{Experiments} \label{sec:experiments}
In this section, we empirically evaluate \texttt{TANGOS} as a regularization method for improving generalization performance. We present our benchmark methods and training architecture, followed by extensive results on real-world datasets. There are a few main aspects that deserve empirical investigation, which we investigate in turn: $\blacktriangleright$ \textbf{Stand-alone performance.} \S5.1 Comparing the performance of \texttt{TANGOS}, where the focus is on applying it as a stand-alone regularizer, to a variety of benchmarks on a suite of real-world datasets. $\blacktriangleright$ \textbf{In tandem performance.} \S5.2 Motivated by our unique regularization objective and our analysis in \S4, we demonstrate that applying \texttt{TANGOS} \emph{in conjunction} with other regularizers can lead to even greater gains in generalization performance. $\blacktriangleright$ \textbf{Modern architectures.} \S5.3 We evaluate performance on a state-of-the-art tabular architecture and compare to boosting. All experiments were run on NVIDIA RTX A4000 GPUs. Code is provided on Github\footnote{\url{https://github.com/alanjeffares/TANGOS}}\footnote{\url{https://github.com/vanderschaarlab/TANGOS}}.

\textbf{TANGOS.} We train \texttt{TANGOS} regularized models as described in \Cref{alg:TANGOS} in Appendix \ref{appndx:compute}. For the specialization parameter we search for $\lambda_1 \in \{1, 10, 100\}$ and for the orthogonalization parameter we search for $\lambda_2 \in \{0.1, 1\}$. For computational efficiency, we apply a sub-sampling scheme where 50 neuron pairs are randomly sampled for each input (for further details see Appendix \ref{appndx:compute}).

\textbf{Benchmarks.} We evaluate \texttt{TANGOS} against a selection of popular regularizer benchmarks. First, we consider weight decay methods \textbf{L1} and \textbf{L2} regularization, which sparsify and shrink the learnable parameters. For the regularizers coefficients, we search for $\lambda \in \{0.1, 0.01, 0.001\}$ where regularization is applied to all layers. 
Next, we consider Dropout (\textbf{DO}), 
with drop rate $p \in \{10\%, 25\%, 50\%\}$, and apply DO after every dense layer during training. We also consider implicit regularization in batch normalization (\textbf{BN}). Lastly, we evaluate data augmentation techniques Input Noise (\textbf{IN}), where we use additive Gaussian noise with mean 0 and standard deviation $\sigma \in \{0.1, 0.05, 0.01\}$ and MixUp (\textbf{MU}). Furthermore, each training run applies early stopping with patience of 30 epochs. In all experiments, we use 5-fold cross-validation to train and validate each benchmark. We select the model which achieves the lowest validation error and provide a final evaluation on a held-out test set.
\vspace{-0.25em}
\subsection{Generalization: Stand-Alone Regularization}
\vspace{-0.25em}
For the first set of experiments, we are interested in investigating the individual regularization effect of \texttt{TANGOS}. To ensure a fair comparison, we evaluate the generalization performance on held-out test sets across a variety of datasets. 

\textbf{Datasets.} We employ $20$ real-world tabular datasets from the UCI machine learning repository. Each dataset is split into $80\%$ for cross-validation and the remaining $20\%$ for testing. The splits are standardized on just the training data, such that features have mean $0$ and standard deviation $1$ and categorical variables are one-hot encoded. See Appendix \ref{appndx:data} for further details on the $20$ datasets used. 

\textbf{Training and Evaluation.} To ensure a fair comparison, all regularizers are applied to an MLP with two ReLU-activated hidden layers, where each hidden layer has $d_H+1$ neurons. The models are trained using Adam optimizer with a dataset-dependent learning rate from $\{0.01, 0.001, 0.0001\}$ and are trained for up to a maximum of $200$ epochs. For regression tasks, we report the average Mean Square Error (MSE) and, on classification tasks, we report the average negative log-likelihood (NLL).

\textbf{Results.} 
Table \ref{tab:stand_alone_generalisation} provides the benchmarking results for individual regularizers. We observe that \texttt{TANGOS} achieves the best performance on $10/20$ of the datasets. We also observe that on $6$ of the remaining datasets, \texttt{TANGOS} ranks second. This is also illustrated by the ranking plot in Appendix~\ref{appdx:ranking_plot}. There we also provide a table displaying standard errors. As several results have overlapping error intervals, we also assess the magnitude of improvement by performing a non-parametric Wilcoxon signed-rank sum test \citep{wilcoxon1992individual} paired at the dataset level.  We compare \texttt{TANGOS} to the best-performing baseline method (L2) as a one-tailed test for both the regression and classification results obtaining p-values of 0.006 and 0.026 respectively. This can be interpreted as strong evidence to suggest the difference is statistically significant in both cases.
Note that a single regularizer is seldom used by itself. In addition to a stand-alone method, it remains to be shown that \texttt{TANGOS} brings value when used with other regularization methods. This is explored in the next section.

\vspace{-0.25em}
\subsection{Generalisation: In Tandem Regularization}
\vspace{-0.25em}
Motivated by the insights described in \S4, a natural next question is whether \texttt{TANGOS} can be applied in conjunction with existing regularization to unlock even greater generalization performance. In this set of experiments, we investigate this question. 

\textbf{Setup.} 
The setting for this experiment is identical to \S5.1 except now we consider the six baseline regularizers \emph{in tandem} with \texttt{TANGOS}. We examine if pairing our proposed regularizer with existing methods results in even greater generalization performance. We again run 5-fold cross-validation, searching over the same hyperparameters, with the final models evaluated on a held-out test set. 

\begin{figure}%
  \vspace{-1em}
    \centering
    \subfloat{{\includegraphics[width=0.45\textwidth]{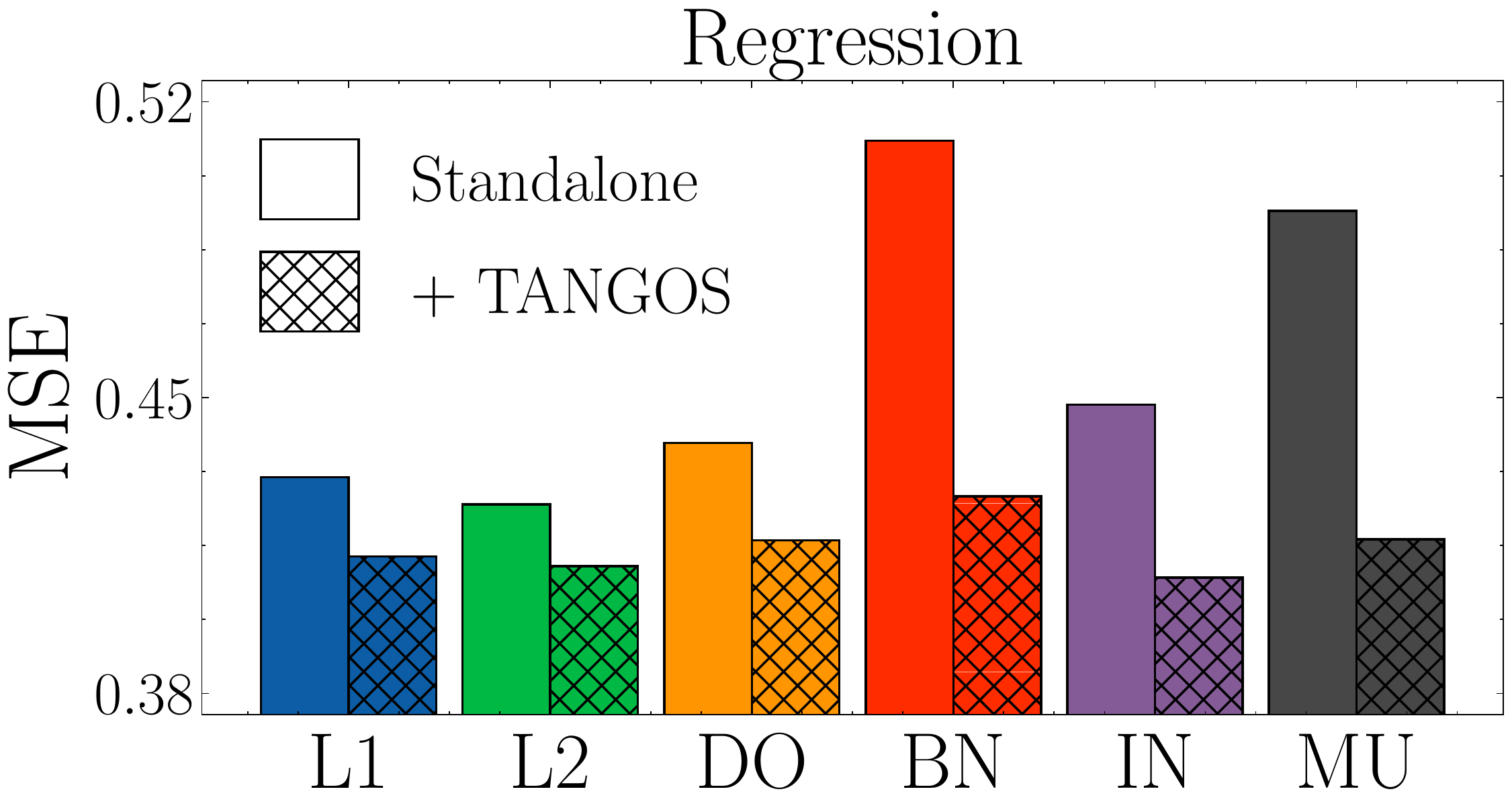} }}%
    \qquad
    \subfloat{{\includegraphics[width=0.45\textwidth]{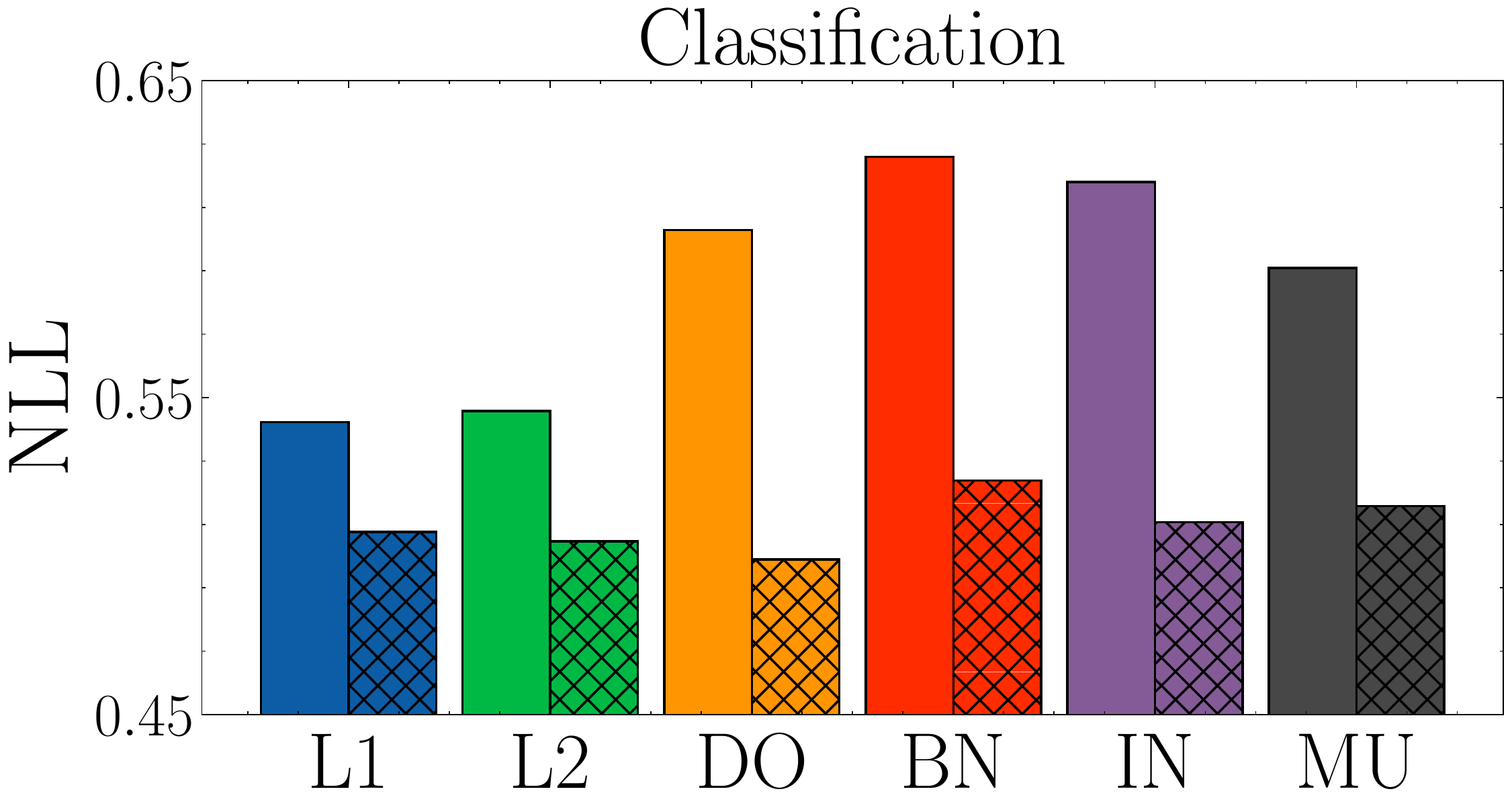} }}%
    \caption{\small \textbf{In Tandem Regularization.} Aggregated errors across the 10 regression datasets (left) and the 10 classification datasets (right). In all cases, the addition of \texttt{TANGOS} provides superior performance over the standalone regularizer.}%
    \label{fig:in-tandem}%
  \vspace{-1em}
\end{figure}

\textbf{Results.}
We summarize the aggregated results over the datasets for each of the six baseline regularizers in combination with \texttt{TANGOS} in Figure \ref{fig:in-tandem}. Consistently across all regularizers in both the regression and the classification settings, we observe that adding \texttt{TANGOS} regularization improves test performance. We provide the full table of results in the supplementary material. We also note an apparent interaction effect between certain regularizers (i.e. input noise for regression and dropout for classification), where methods that seemed to not be particularly effective as stand-alone regularizers become the best-performing method when evaluated in tandem. The relationship between such regularizers provides an interesting direction for future work.

\subsection{Closing the Gap on Boosting}
In this experiment, we apply \texttt{TANGOS} regularization to a state-of-the-art deep learning architecture for tabular data \citep{gorishniy2021revisiting} and evaluate its contribution towards producing competitive performance against leading boosting methods. We provide an extended description of this experiment in \Cref{app:boosting} and results in \Cref{tab:boosting}. We find that \texttt{TANGOS} provides moderate gains in this setting, improving performance relative to state-of-the-art boosting methods. Although boosting approaches still match or outperform deep learning in this setting, in \Cref{app:motivation} we argue that deep learning may also be worth pursuing in the tabular modality for its other distinct advantages. 
\begin{table}[h!]
\caption{\small \textbf{FT-Transformer Architecture and Boosting.} Adding \texttt{TANGOS} regularization can contribute to closing the gap between state-of-the-art tabular architectures and leading boosting methods. We report mean accuracy $\pm{}$ standard deviation.}
\label{tab:boosting}
\centering
\midsepremove
\resizebox{0.9\linewidth}{!}{
\begin{tabular}{|c|c|cc|cc|}
\toprule
 \multirow{ 2}{*}{Setting}&
 \multirow{ 2}{*}{Dataset}& \multicolumn{2}{|c|}{\cellcolor[HTML]{EFEFEF}FT-Transformer} & \multicolumn{2}{|c|}{\cellcolor[HTML]{EFEFEF}Boosting} \\ \cline{3-6}
& & Baseline & + \texttt{TANGOS} & XGBoost & CatBoost \\ \midrule
\multirow{ 2}{*}{Default} & Jannis & $0.714 \pm{0.002}$ & $\mathbf{0.720} \pm{0.000}$ & $0.711 \pm{0.000}$ & $\mathbf{0.724}\pm{0.001}$  \\
 & Higgs & $0.721 \pm{0.002}$ & $\mathbf{0.723} \pm{0.000}$ & $0.717 \pm{0.000}$ & $\mathbf{0.728}\pm{0.001}$ \\
 \hline
\multirow{ 2}{*}{Tuned} & Jannis & $0.720 \pm{0.001}$ & $\mathbf{0.727} \pm{0.001}$ & $0.724 \pm{0.000}$ & $\mathbf{0.727}\pm{0.001}$  \\
 & Higgs & $0.727 \pm{0.002}$ & $\mathbf{0.729} \pm{0.002}$ & $0.728 \pm{0.001}$ & $\mathbf{0.729}\pm{0.002}$ \\

\bottomrule
\end{tabular}
}
\end{table}

\vspace{-0.2em}
\section{Discussion}
\vspace{-0.2em}
In this work, we have introduced \texttt{TANGOS}, a novel regularization method that promotes specialization and orthogonalization among the gradient attributions of the latent units of a neural network. 
We showed \emph{how} this regularization objective is distinct from other popular methods and motivated \emph{why} it provides out-of-sample generalization. We empirically demonstrated \texttt{TANGOS} utility with extensive experiments.
This work raises several exciting avenues for \textbf{future work} including (1) developing \texttt{TANGOS} beyond the tabular setting (e.g. images), (2) investigating alternative efficient methods for achieving specialization and orthogonalization, (3) proposing other latent gradient attribution regularizers, (4) augmenting \texttt{TANGOS} for specific applications such as multi-modal learning or increased interpretability (see \Cref{app:motivation}).

\newpage

\section*{Acknowledgments}
We thank the anonymous ICLR reviewers as well as members of the van der Schaar lab for many insightful comments and suggestions. Alan Jeffares is funded by the Cystic Fibrosis Trust. Tennison Liu would like to thank AstraZeneca for their sponsorship and support.
Fergus Imrie and Mihaela van der Schaar are supported by the National Science Foundation (NSF, grant number 1722516). Mihaela van der Schaar is additionally supported by the Office of Naval Research (ONR).

\section*{Reproducibility Statement}
We have attempted to make our experimental results easily reproducible by both a detailed description of our experimental procedure and providing the code used to produce our results (\url{https://github.com/alanjeffares/TANGOS}). Experiments are described in Section \ref{sec:experiments} with further details in Appendices \ref{appndx:compute} and \ref{appndx:data}. All datasets used in this work can be freely downloaded from the UCI repository \citep{dua2017uci} with specific details provided in Appendix \ref{appndx:data}.

\bibliography{camera_ready.bib}
\bibliographystyle{iclr2023_conference.bst}

\newpage
\appendix

\section{Extended Related Works} \label{app:related-works}
\textbf{Neural Network Regularization.} Regularization methods seek to penalize complexity and impose a form of smoothness on a model. This may be cast as expressing a prior belief over the hypothesis space of a neural network which attempts to aid generalization. $\blacktriangleright$ \textbf{Categories.} A vast array of regularization methods have been proposed throughout the literature (for a comprehensive taxonomy see e.g. \citealp{kukavcka2017regularization}). Modern nomenclature typically includes broad modeling decisions such as choice of architecture, loss function, and optimization method under the umbrella of regularization. Additionally, many regularization techniques augment the training data using methods such as input noise \citep{krizhevsky2012imagenet} or MixUp \citep{zhang2017mixup}. Dropout \citep{hinton2012improving} and related approaches that augment a hidden representation of the input may also be included in this category. Possibly a more conventional category of regularization is that which adds explicit penalty term(s) to the loss function. These terms might penalize the network weights directly to shrink or sparsify their values as in L2 \citep{hoerl1970ridge} and L1 \citep{tibshirani1996regression}, respectively. Alternatively, network outputs may be penalized to, for example, reduce overconfidence \citep{pereyra2017regularizing}. 
$\blacktriangleright$~\textbf{Weight Orthogonalization.} A number of works have studied the orthogonalization of network weights via various weight penalization methods \citep{bansal2018can}. More recent work in \cite{liu2021orthogonal} proposed to learn an orthogonal transformation of the randomly initialized incoming weights to a given neuron. In contrast, this work seeks to ensure that the gradients of different latent neurons with respect to a given input vector are orthogonal.
$\blacktriangleright$~\textbf{Combination.} Compositions of multiple regularization methods are extensively applied in practice. An early example in the regression setting is the elastic net penalty \citep{zou2005regularization} which attempts to combine sparsity with shrinkage in the coefficients. More recent work has demonstrated the effectiveness of combining several regularization terms on tabular data \citep{kadra2021well}, a domain in which neural networks superiority had previously been less convincing. 

\section{Tabular Architectures and Boosting}\label{app:boosting}
While non-neural methods such as XGBoost \citep{chen2016xgboost} and CatBoost \citep{prokhorenkova2018catboost} are still considered state of the art for tabular data \citep{grinsztajn2022tree}, much progress has been made in recent years to close the gap. Furthermore, differing learning paradigms have various strengths and weaknesses outside of maximum generalization performance, which is often a consideration in practical applications. While boosting methods boast excellent computational efficiency and strong out-of-the-box performance, neural networks have unique utility in, for example, multi-modal learning \citep{ramachandram2017deep}, meta-learning \citep{hospedales2021meta} and certain interpretability methods \citep{zhang2021survey}.
In this section, we provide additional experiments applying \texttt{TANGOS} to a state-of-the-art transformer architecture for tabular data proposed in \cite{gorishniy2021revisiting}. Specifically, this architecture combines a Feature Tokenizer which transforms features into embeddings with a multi-layer Transformer \citep{vaswani2017attention}. We compare this FT-Transformer architecture to boosting methods in the \textit{default} setting where we evaluate out-of-the-box performance and the \textit{tuned} setting where we jointly optimize the Transformer along with its baseline regularizers. We describe these two settings in more detail next.

\textbf{Default Setting.} In this setting, we use a 3-layer Transformer with a 32-dimensional feature embedding size and 4 attention heads. Following the original paper we use Reglu activations, a hidden layer size of 43 corresponding to a ratio of $\frac{4}{3}$ with the embedding size, Kaiming initialization \citep{he2015delving}, and AdamW optimizer \citep{loshchilov2017decoupled}. Finally, we apply a learning rate of 0.001. We compare this architecture with and without \texttt{TANGOS} regularization applied which we refer to as ``Baseline'' and ``+ \texttt{TANGOS}'' respectively. We set $\lambda_1 = 1$ and $\lambda_2 = 0.01$ which were found to be reasonable default values for specialization and orthogonalization in our experiments in Section \ref{sec:experiments}.

\textbf{Tuned Setting.} Here we apply ten iterations of random search tuning over the same hyperparameters as in the original work with those achieving the best validation performance selected. We then evaluate this combination by training over three seeds and perform their final evaluations on a held-out test set. We search using the same distributions as in the original work and consider the following ranges. L2 regularization $\in [1e-06, 1e-03]$, residual dropout $\in [0.0, 0.2]$, hidden layer dropout $\in [0.0, 0.5]$, attention dropout $\in [0.0, 0.5]$, hidden layer to feature embedding dimension ratio $\in [1.0, 3.0]$, embedding dimension $\in [16, 48]$, number of layers $\in [1, 3]$, learning rate $\in [1e-04, 1e-03]$. In the ``+ \texttt{TANGOS}'' setting we also include $\lambda_1 \in [0.001, 10]$ and $\lambda_2 \in [0.0001, 1]$ with a log uniform distribution. All remaining architecture choices are consistent with the default setting and the original work.

We ran our experiments on the Jannis \citep{guyon2019analysis} and Higgs \citep{baldi2014searching} datasets. These are both classification datasets consisting of $83733$ and $98050$ examples respectively. These datasets were selected as they represent a significant number of input examples along with a middling number of input features relative to the other tabular datasets explored in this work ($54$ and $28$ respectively). We follow the experimental protocol of the boosting comparison in \cite{grinsztajn2022tree} using the same training, validation, and test splits and reporting mean test accuracy over three runs. Therefore we obtain the same results for boosting as reported in that work.

The results of this experiment are reported in Table \ref{tab:boosting} where we find that \texttt{TANGOS} does indeed have a positive effect on the FT-Transformer performance although, consistent with the original work, we found that regularization only provides modest gains at best with this architecture. While we do not claim that \texttt{TANGOS} regularization results in neural networks that outperform Boosting methods, these results indicate that \texttt{TANGOS} regularization can contribute to closing the gap and may play a key role when combined with other methods as highlighted in \cite{kadra2021well}. We believe this to be an important area for future research and, in particular, expect that architecture-specific developments of the ideas presented in this work may provide further improvements on the results obtained in this section.

\section{Motivation for Deep Learning on Tabular Data} \label{app:motivation}

Several works have argued that boosting methods generally achieve superior performance to even state-of-the-art deep learning architectures for tabular data \citep{grinsztajn2022tree, shwartz2022tabular}. However, this is in contrast to recent findings for transformer style architectures in \cite{gorishniy2021revisiting}, especially with appropriate feature embeddings \citep{gorishniy2022embeddings} and sufficient pretraining \citep{rubachev2022revisiting}. We defer from this discussion to highlight a selection of reasons to consider deep learning methods for tabular data \textit{beyond} straightforward improvements in predictive performance. In particular, we include a number of deep learning paradigms that are difficult to analogize for non-neural models and have been successfully applied to tabular data.

\textbf{Multi-modal learning} refers to the task of modeling data inputs that consist of multiple data modalities (e.g. image, text, tabular). As one might intuit, jointly modeling these multiple modalities can result in better performance than independently predicting from each of them \citep{ramachandram2017deep, guo2019deep}. Deep learning provides a uniquely natural method of combining modalities with the advantages of (1) modality-specific encoders, (2) that are fused into a joint downstream representation and trained end-to-end with backpropagation, and (3) superior modeling performance in many modalities such as images and natural language. Healthcare is a domain in which multi-modal learning is particularly salient \citep{acosta2022multimodal}. Recent work in \cite{wu2022deep} showed that jointly modeling tabular clinical records using an MLP together with medical images using a CNN outperforms the non-multi-modal baselines. Elsewhere in \cite{tang2020deep}, a multi-modal approach is taken in combining input modalities based on the preprocessing of functional magnetic resonance imaging and region of interest time series data for the diagnosis of autism spectrum disorder. A resnet-18 encodes one modality while an MLP encodes the other, resulting in superior performance when analyzed in an ablation study. In this setting, progress in modeling each of the individual modalities is likely to result in better performance of the system as a whole. Interestingly, \cite{ramachandram2017deep} identified regularization techniques for improved cross-modality learning as an important research direction. We believe that further development of the ideas presented in this work could provide a powerful tool for balancing how models attend to multiple input modalities.

\textbf{Meta-learning} aims to distill the experience of multiple learning episodes across a distribution of related tasks to improve learning performance on future tasks \citep{hospedales2021meta}. Deep learning-based approaches have seen great success as a solution to this problem in a variety of fields. In the tabular domain, with careful consideration of the shared information between tasks, recent works have also shown promising results in this direction by developing methods for transferring deep tabular models across tables \citep{wang2022transtab,levin2022transfer}. In particular, in \cite{levin2022transfer} it was noted that ``representation learning with deep tabular models provides significant gains over strong GBDT baselines'', also finding that ``the gains are especially pronounced in low data regimes''.

\textbf{Interpretability} is an important area of deep learning research aiming to provide users with the ability to understand and reason about model outputs. Certain classes of interpretability methods have recently been developed that provide distinct forms of interpretability relying on the hidden representations of neural networks. In such models, probing the representation space of a deep model permits a new type of interpretation. For instance, \cite{kim2018interpretability} studies how human concepts are represented by deep classifiers. This makes it possible to analyze how the classes predicted by the model relate to human understandable concepts. For example, one can verify if the stripe concept is relevant for a CNN classifier to identify a zebra, as demonstrated in the paper. Another example is \cite{crabbe2021explaining}, which proposes to explain a given example with reference to a freely selected set of other examples (potentially from the same dataset). A user study was carried out in this work which concluded that, among non-technical users, this method of explanation does affect their confidence in the model’s prediction. These powerful methods crucially rely on the model's representation space, which effectively assumes that the model is a deep neural network.

\textbf{Representation learning} more generally provides access to several other methods from deep learning to the tabular domain. A number of works have used deep learning approaches to map inputs to embeddings which can be useful for downstream applications. SuperTML \citep{sun2019supertml} and \cite{zhu2021converting} map tabular inputs to image-like embeddings that can therefore be passed to image architectures such as CNNs. Other self-supervised methods include VIME \citep{yoon2020vime} which applies input reconstruction, SubTab \citep{ucar2021subtab} which suggests a multi-view reconstruction task and SCARF \citep{bahri2021scarf} which takes a contrastive approach. Representation learning approaches such as these have proven successful on downstream tabular data tasks such as uncertainty quantification \citep{seedat2023improving}, federated learning \citep{he2022hybrid}, anomaly detection \citep{liang2022self}, and feature selection \citep{lee2022self}.

\section{TANGOS Behavior Analysis}
In this section, we apply \texttt{TANGOS} to a simple image classification task using a convolutional neural network (CNN) and provide a qualitative analysis of the behavior of the learned network. This analysis is conducted on the MNIST dataset \citep{lecun1998gradient} using the recommended split resulting in 60,000 training and 10,000 validation examples.

In this experiment, we train a standard CNN architecture (as described in Table \ref{tab:architecture}) with a penultimate hidden layer of 10 neurons for 10 epochs with Adam optimizer and a learning rate of 0.001. We also apply L2 regularization with weight 0.001. After each epoch, the model is evaluated on the validation set where the epoch achieving the best validation performance is stored for further analysis. Two models are trained under this protocol. One model which applied \texttt{TANGOS} to the penultimate hidden layer with $\lambda_1 = 100$, $\lambda_2 = 0.1$ and $M = 25$ and a baseline model which does not apply \texttt{TANGOS}.

\begin{table}[h]
    \vspace{-0.5em}
	\setstretch{1.5}
	\caption{MNIST Convolutional Neural Network Architecture.}
	\label{tab:architecture}
	\begin{center}
		\begin{adjustbox}{width=\columnwidth}
			\begin{tabular}{clc} 
				\toprule
				Layer Type & Hyperparameters & Activation Function\\
				\hline
                Conv2d & \makecell[l]{Input Channels:1 ; Output Channels:16 ; Kernel Size:5 ; Stride:2 ; Padding:1 } & ReLU  \\ 
				Conv2d & \makecell[l]{Input Channels:16 ; Output Channels:32 ; Kernel Size:5 ; Stride:2 ; Padding:1 } & ReLU \\ 
				Flatten & Start Dimension:1 &   \\ 
				Linear & Input Dimension: 512 ; Output Dimension: 256 & ReLU  \\
				Linear & Input Dimension: 256 ; Output Dimension: 10 &  \\ 
				Linear & Input Dimension: 10 ; Output Dimension: 10 &  \\ 
				\bottomrule
			\end{tabular}
		\end{adjustbox}
	\end{center}
\end{table}
\vspace{-0.25em}

\begin{figure}[!htb]
  \vspace{-0.5em}
  \centering
  \includegraphics[width=0.9\linewidth]{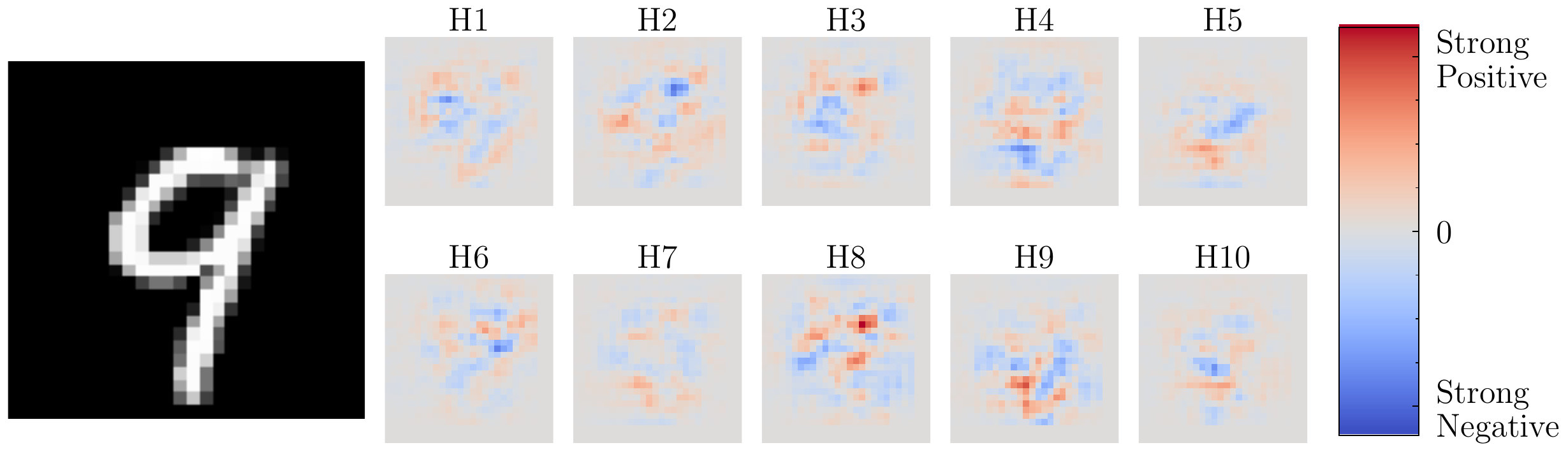}
  \caption{\textbf{Without TANGOS Training.} Gradient attributions with respect to each of the 10 hidden neurons. These results suggest significant overlap among the gradient attributions.}
  \label{fig:reg_saliency}
  \vspace{-0.5em}
\end{figure}

\begin{figure}[!htb]
  \vspace{-0.5em}
  \centering
  \includegraphics[width=0.9\linewidth]{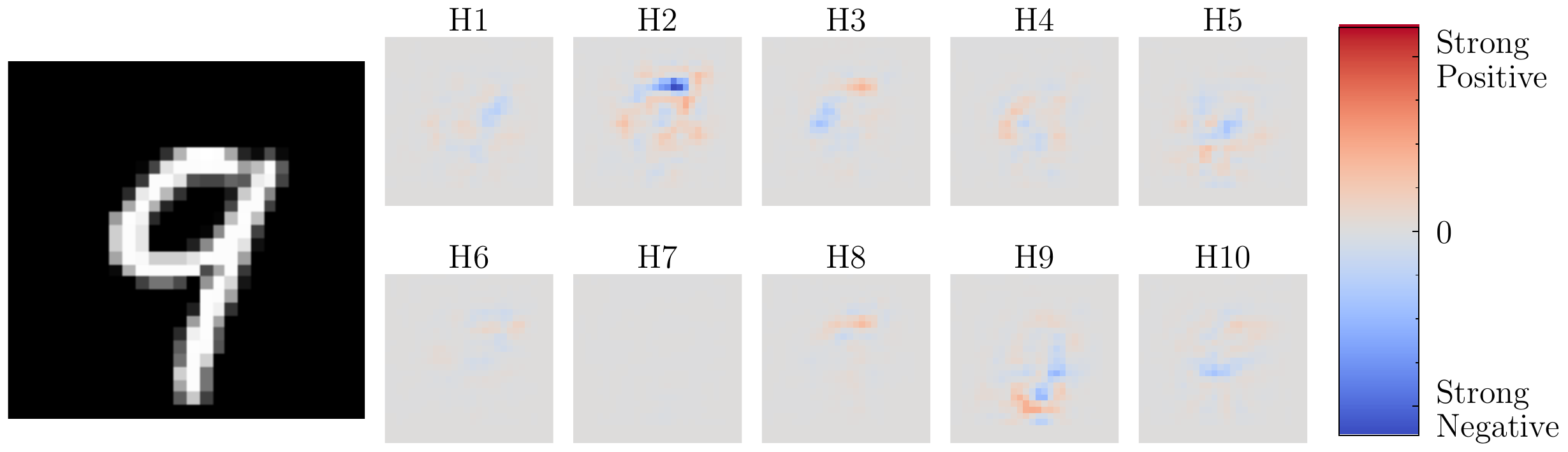}
  \caption{\textbf{With TANGOS Training.} \texttt{TANGOS} encourages gradient attributions to be sparse with minimal overlap.}
  \label{fig:LGOS_saliency}
  \vspace{-0.5em}
\end{figure}

In this section, we examine the gradients of each of the 10 neurons in the penultimate hidden layer with respect to each of the input dimensions of a given image. \texttt{TANGOS} is designed to reward orthogonalization and specialization of these gradient attributions which can be evaluated qualitatively by inspection. In all plots that follow we apply a min-max scaling across all hidden units for a fair comparison. Both strong positive and strong negative values for attributions may be interpreted as a latent unit attending to a given input dimension. In Figure \ref{fig:reg_saliency} we provide results for the baseline model applied to a test image where, in line with similar analyses in previous works such as \cite{crabbe2022label}, we note that the way in which hidden units attend to the input is highly entangled. In contrast to this, in Figure \ref{fig:LGOS_saliency}, we include the same plot for the \texttt{TANGOS} trained model on the same image. In this case, each hidden unit does indeed produce relatively sparse and orthogonal attributions as desired. These results were consistent across the test images.

\begin{figure}[!htb]
  \centering
  \includegraphics[width=0.9\linewidth]{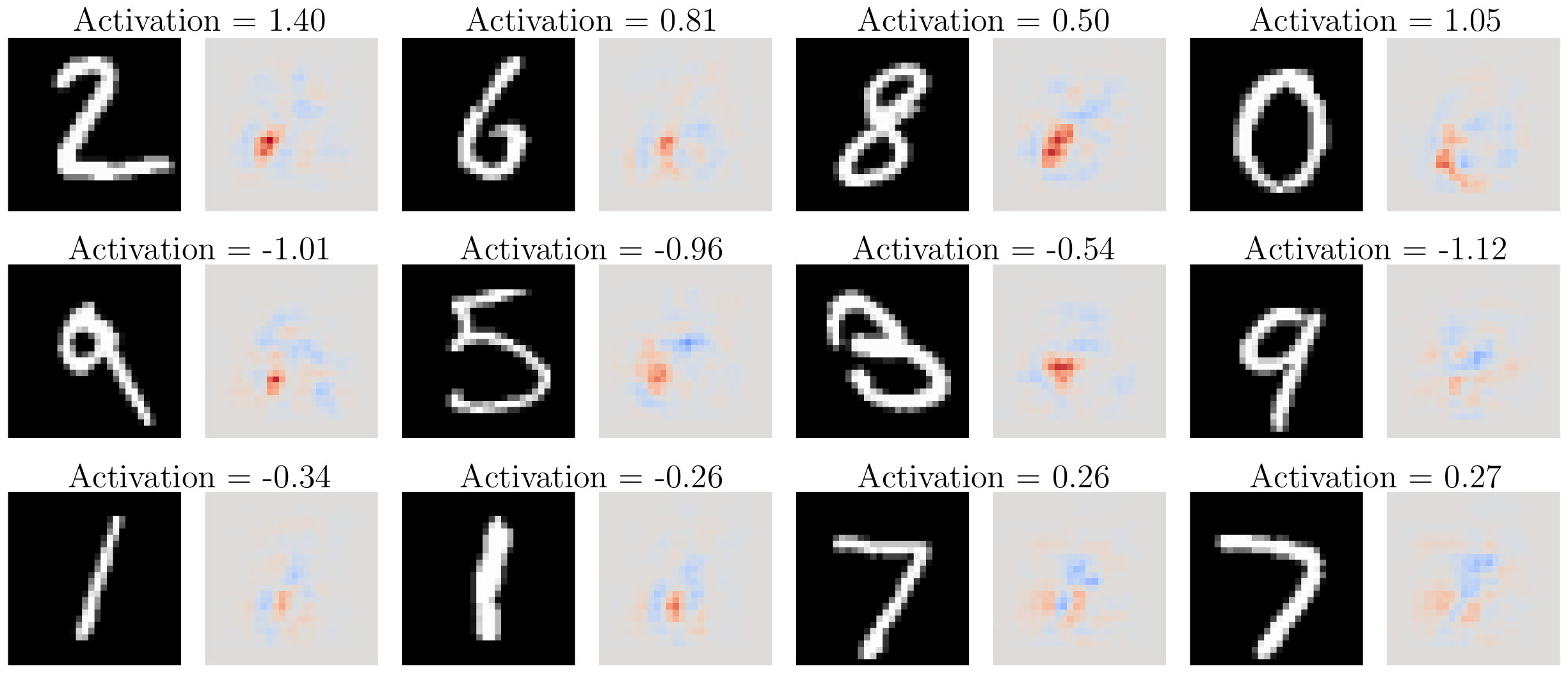}
  \caption{\textbf{Hidden Neuron 5.} This neuron attempts to discriminate whether inputs contain a closed loop on the lower left of their digit. Inputs with a closed lower loop incur highly positive activations (first row). Inputs with open lower loops incur highly negative activations (second row). While ambiguous inputs with no lower loop at all tend to produce low-magnitude activations (third row).}
  \label{fig:neuron_5}
  \vspace{-0.5em}
\end{figure}

We can glean further insight into the \texttt{TANGOS} trained model by examining the role of individual neurons across multiple test images. In Figure \ref{fig:neuron_5}, we provide the gradient attributions for hidden neuron 5 (H5) from our previous discussion across twelve test images. This neuron appears to discriminate between an open or a closed loop at the lower left of the digit. Indeed this is a key aspect of distinction between the set of digits $\{ 2,6,8,0\}$ (first row) and $\{ 9,5,3\}$ (second row). We also include digits where this visual feature is less useful as they contain no lower-left loop either open or closed (third row). This hypothesis can be further examined by analyzing the values of these activations. We note that the first two rows typically have higher magnitude with opposite signs while the third row has lower magnitude activations. In Table \ref{tab:neuron_5} we summarize the effect of these activation scores on class probabilities by accounting for the weights connecting to each of the ten classes. As one might expect, the weight connections between the hidden neuron and classes on the first row and the second row have opposite signs indicating that neuron 5 does indeed discriminate between these classes.

\begin{table}[h!]
\centering
\caption{\textbf{Neuron 5 Class Weights.} Weights connecting neuron 5 to each of the ten classes and a summary of their combined effect with the neuron activation. The classification influence column provides a categorical indication of the magnitude of the contribution to each class output for a fixed activation magnitude. This is determined by the magnitude of the connecting weight where: Low $\in [0, 0.59]$, Medium $\in [0.59, 1.17]$, and High $\in [1.17, 1.76]$.}
\label{tab:neuron_5}
\vspace{0.1in}
\resizebox{0.8\textwidth}{!}{%
\begin{tabular}{c c c c } 
\hline
\multicolumn{1}{c}{Class label} & \multicolumn{1}{c}{Weight connection} & \multicolumn{1}{c}{\begin{tabular}{c}
  Increases class probability \\ if activation is
\end{tabular}} & \multicolumn{1}{c}{Classification influence}  \\ \hline
0 & 1.0883 & Positive & Medium \\
1 & -0.6826 & Negative & Medium  \\
2 & 1.5298 & Positive & High \\
3 & -1.5862 & Negative & High  \\
4 & -0.2516 & Negative & Low  \\
5 & -0.6008 & Negative & Medium  \\
6 & 0.9065 & Positive & Medium  \\
7 & 0.3524 & Positive & Low  \\
8 & 0.7362 & Positive & Medium  \\
9 & -1.7608 & Negative & High  \\

\hline
\end{tabular}}
\end{table}

\section{Performance with Increasing Data Size}
In this section, we evaluate \texttt{TANGOS} performance with an increasing number of input examples. To do this we use the Dionis dataset, which was the largest benchmark dataset proposed in \cite{kadra2021well} with 416,188 examples. As in that work, we set aside 20\% for testing with the remaining data further split into 80\% training and 20\% validation. The data was standardized to have zero mean and unit variance with statistics calculated on the training data. We then consider using various proportions (10\%, 50\%, 100\%) of the training data to train an MLP with and without \texttt{TANGOS} regularization. We also evaluate the best-performing regularization method, L2, from our experiments in Section \ref{sec:experiments}. For both regularization methods, we train three hyperperameter settings at each proportion and evaluate the best performing of the three on the test set. For TANGOS we consider $\{(\lambda_1=1, \lambda_2 = 0.01), (\lambda_1=1, \lambda_2 = 0.1), (\lambda_1=10, \lambda_2 = 0.1)\}$ and for L2 we consider $\lambda \in \{0.01, 0.001, 0.0001 \}$. We repeat this procedure for 6 runs and report the mean test accuracy. The MLP contained three ReLU-activated hidden layers of 400, 100, and 10 hidden units, respectively. 

\begin{figure}[!ht]
  \vspace{-0.5em}
  \centering
  \includegraphics[width=0.8\linewidth]{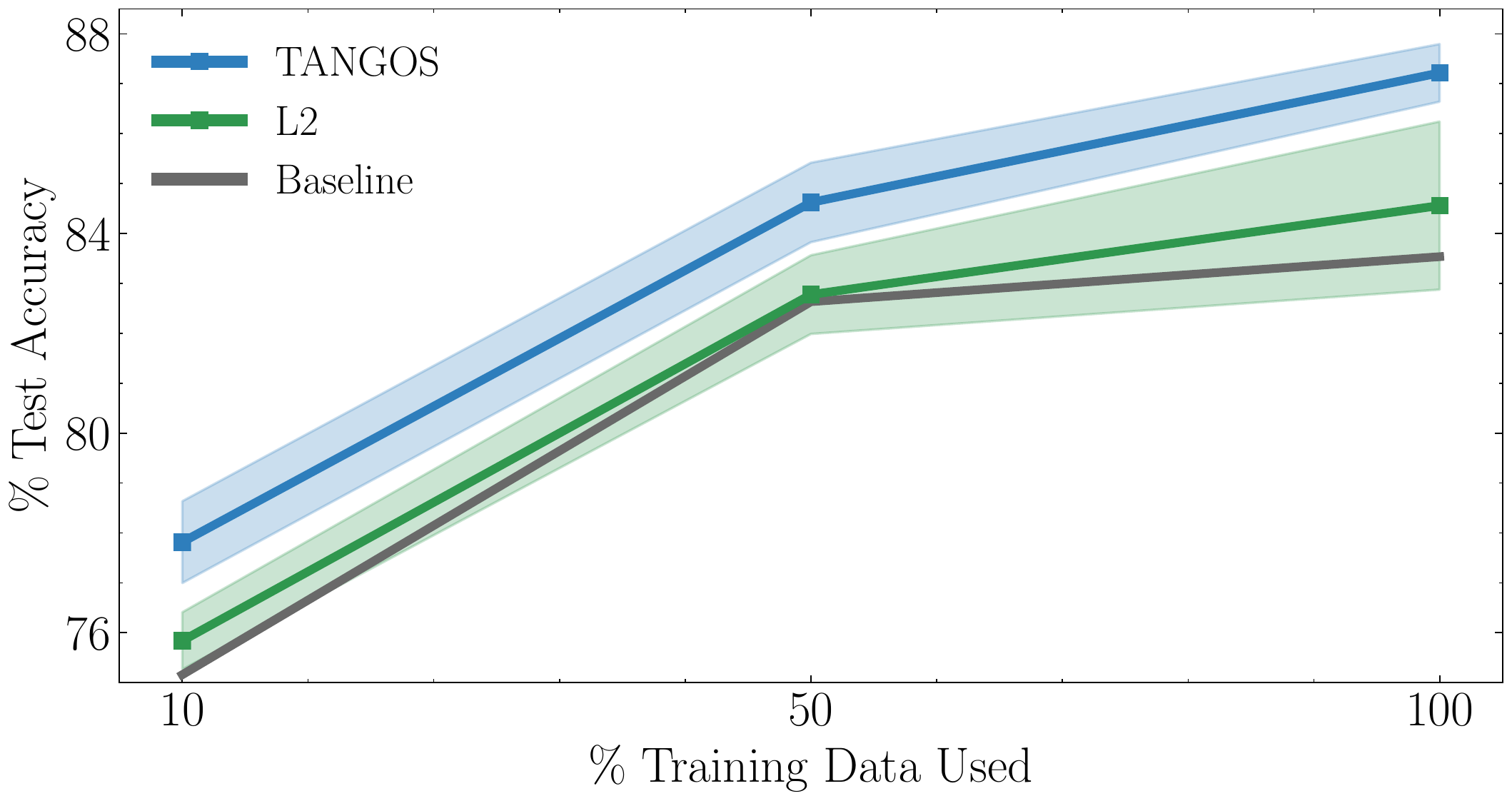}
  \caption{\textbf{Performance Gains With Increasing Data Size.} Training with various proportions of training data from the 416,188 examples of the Dionis dataset, we find the relative boost in performance from \texttt{TANGOS} to be consistent.}
  \label{fig:scale_performance}
  \vspace{-0.5em}
\end{figure}

We include the results of this experiment in Figure \ref{fig:scale_performance}. Consistent with our experiments in Section \ref{sec:experiments}, we find that \texttt{TANGOS} outperforms both the baseline model and the strongest baseline regularization method across all proportions of the data. These results are indicative that \texttt{TANGOS} remains similarly effective across both small and large datasets in the tabular domain.

\section{Approximation and Algorithm} \label{appndx:compute}
Calculating the attribution of the latent units with respect to the input involves computing the Jacobian matrix, which can be computed in $\mathcal{O}(1)$ time and has memory complexity $\mathcal{O}(d_Hd_X)$. The computational complexity of calculating $\mathcal{L}_{orth}$ is $\mathcal{O}(d_H^2)$ (i.e. all pairwise computation between latent units). While the calculation can be efficiently parallelized, this still becomes impractically expensive with higher dimensional layers. To address this, we introduce a relaxation by randomly subsampling pairs of neurons to calculate attribution similarity. We denote by $I$ denote the set of all possible pairs of neuron indices, $I = \{(i, j) \:\forall\: i, j \in [d_H] \:\text{and}\: i\neq j\}$. Further, we let $M$ denote a randomly sampled subset of $I$, $M \subseteq I$. We devise an approximation to the regularization term, denoted by $\mathcal{L}^{\prime}_{orth}$, by estimating the penalty on the subset $M$, where the size of $M$ can be chosen to balance computational burden with more faithful estimation:
\begin{equation*}
    \mathcal{L}^{\prime}_{orth}(x) = \frac{1}{B} \sum_{b=1}^B\frac{1}{|M|}\sum_{(i, j)\in M}\rho[a^i(x_b), a^j(x_b)]
\end{equation*}

This reduces the complexity of calculating $\mathcal{L}_{orth}$ from $\mathcal{O}(d_H^2)$ to $\mathcal{O}(M)$. For our experimental results described in \Cref{tab:stand_alone_generalisation,tab:stand_alone_uncertainty,tab:paired_results}, we use $|M|=50$. We empirically demonstrate that this approximation still leads to strong results in real-world experiments. The overall training procedure is described in \Cref{alg:TANGOS}.

\begin{algorithm}
\caption{\texttt{TANGOS} regularization}
\label{alg:TANGOS}
\begin{algorithmic}
\item \textbf{Result: } Learned parameters $\theta$
\item \textbf{Input: } $\lambda_1, \lambda_2, \textrm{training data } \mathcal{D}, \textrm{learning rate } \eta$;
\item Initialise $\theta$;
\While{not converged}
        \State Sample $\mathcal{D}_{mini}$ from $\mathcal{D}$;
        \State $\hat{\mathcal{L}}(f_\theta(x), y) = \mathbb{E}_{(x, y)\sim\mathcal{D}_{mini}}[\mathcal{L}(f_\theta(x), y)]$;
        \State $\hat{\mathcal{R}}(x) = \lambda_1\mathbb{E}_{x\sim\mathcal{D}_{mini}}[\mathcal{L}_{spec}(x)] + \lambda_2\mathbb{E}_{x\sim\mathcal{D}_{mini}}[\mathcal{L}_{orth}^\prime(x)]$;
        \State $\theta \leftarrow \theta + \eta\nabla_\theta\left[\hat{\mathcal{L}}(f_\theta(x), y)+\hat{\mathcal{R}}(x)\right]$;
      \EndWhile
\end{algorithmic}
\end{algorithm}

Additionally, we provide an empirical analysis of \texttt{TANGOS} designed to evaluate the effectiveness of our proposed subsampling approximation with respect to generalization performance and computational efficiency as the number of sampled neuron pairs $M$ grows. Furthermore, we analyze the computational efficiency of our method as the number of latent units grows, evaluating the method's capacity to scale to large models.

All experiments are run on the BC dataset which we split into 80\% training and 20\% validation. We fix $\lambda_1 = 100$ and $\lambda_2 = 0.1$ throughout our experiments. We run each experiment over 10 random seeds and report the mean and standard deviation. All remaining experimental details are consistent with our experiments in Section \ref{sec:experiments}. We note that our implementation of \texttt{TANGOS} is not optimized to the same extent as the Pytorch \citep{paszke2019pytorch} implementation of L2 to which we compare, and therefore we may consider our relative computational performance to be a loose upper bound on a truly optimized version. 

\begin{figure}%
    \vspace{-0.1cm}
    \centering
    \subfloat{{\includegraphics[width=0.45\textwidth]{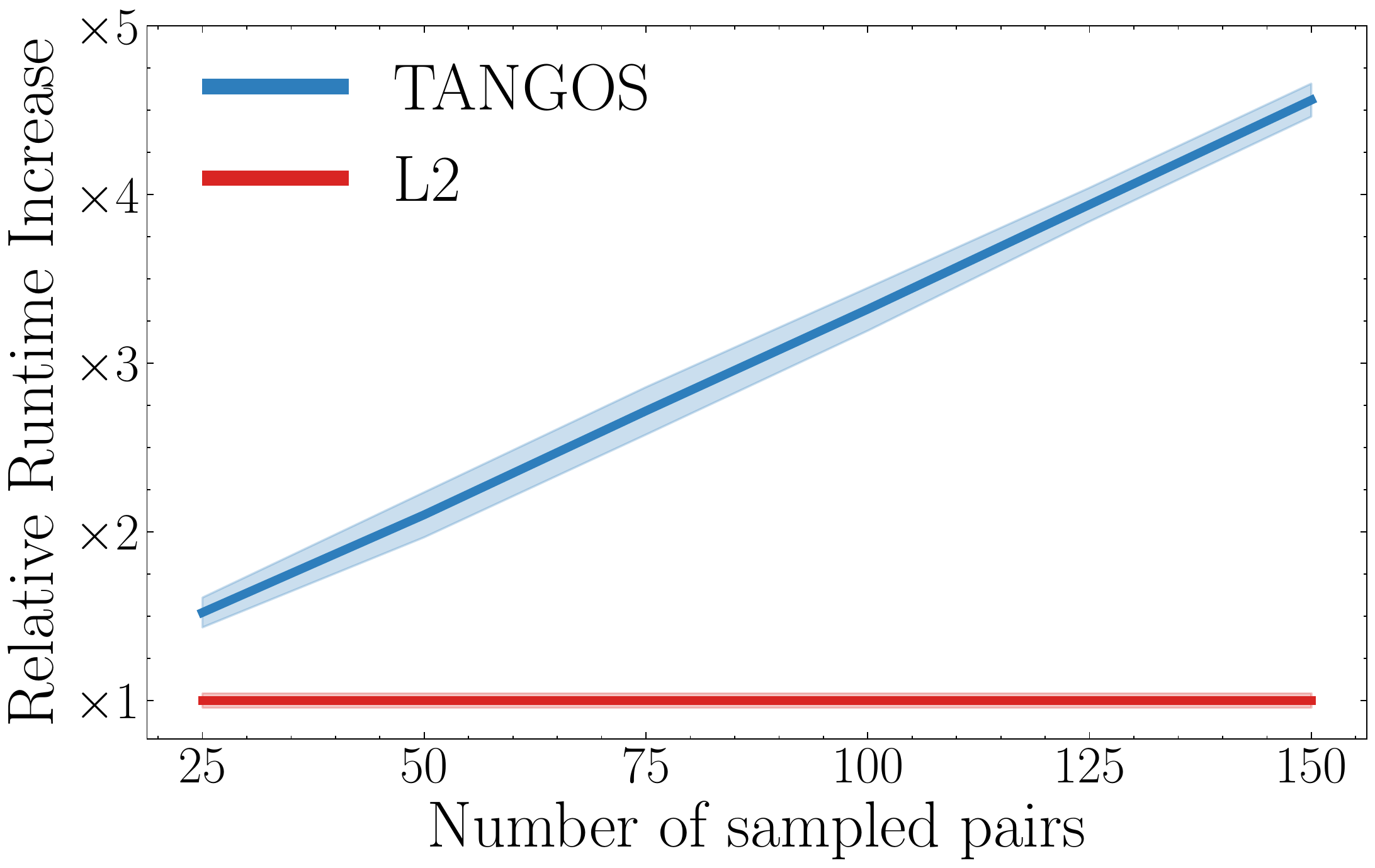} }}%
    \qquad
    \subfloat{{\includegraphics[width=0.45\textwidth]{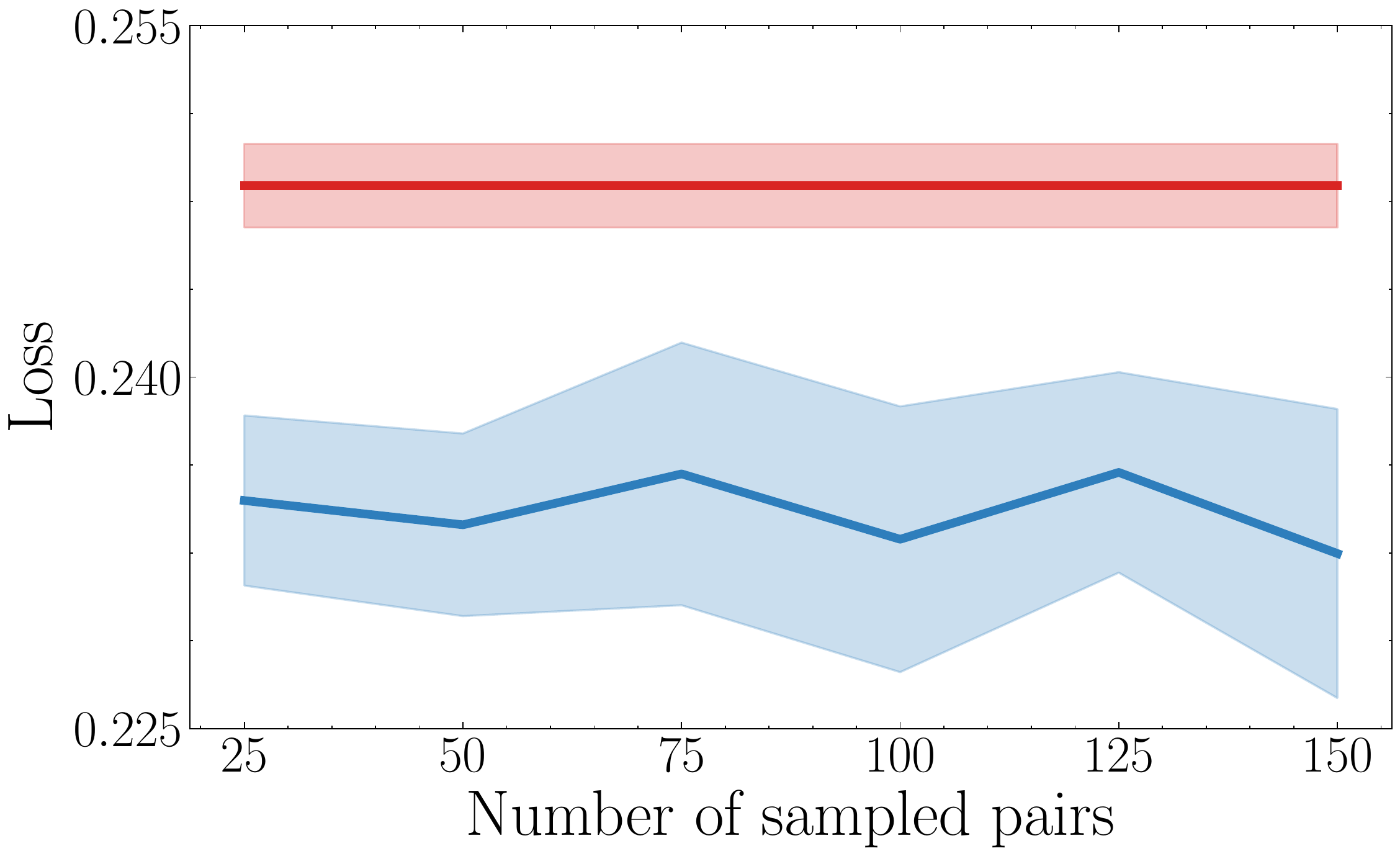} }}%
    \caption{\textbf{Sampling efficiency.} Runtime increases linearly with the number of sampled pairs (left) while better generalization performance is maintained even for low sampling rates (right). The benefits of \texttt{TANGOS} can be realized using our proposed sampling approximation with comparable runtime to even the most efficient existing regularization approaches.}%
    \label{fig:compute}%
    \vspace{-0.1cm}
\end{figure}

In Figure \ref{fig:compute} (left), we report the relative increase in compute time per epoch as we increase the number of sampled pairs. As theory would suggest, this growth is linear. A natural follow-up question is the extent to which model performance is affected by decreasing the number of sampled pairs. In Figure  \ref{fig:compute} (right), we observe that even very low sampling rates still result in excellent performance. Based on these results, our recommendation for practitioners is that while increasing the sampling rate can lead to marginal improvements in performance, relatively low sampling rates appear to be generally sufficient and do not require prohibitive computational overhead. 

\begin{figure}[!htb]
  \centering
  \includegraphics[width=0.65\linewidth]{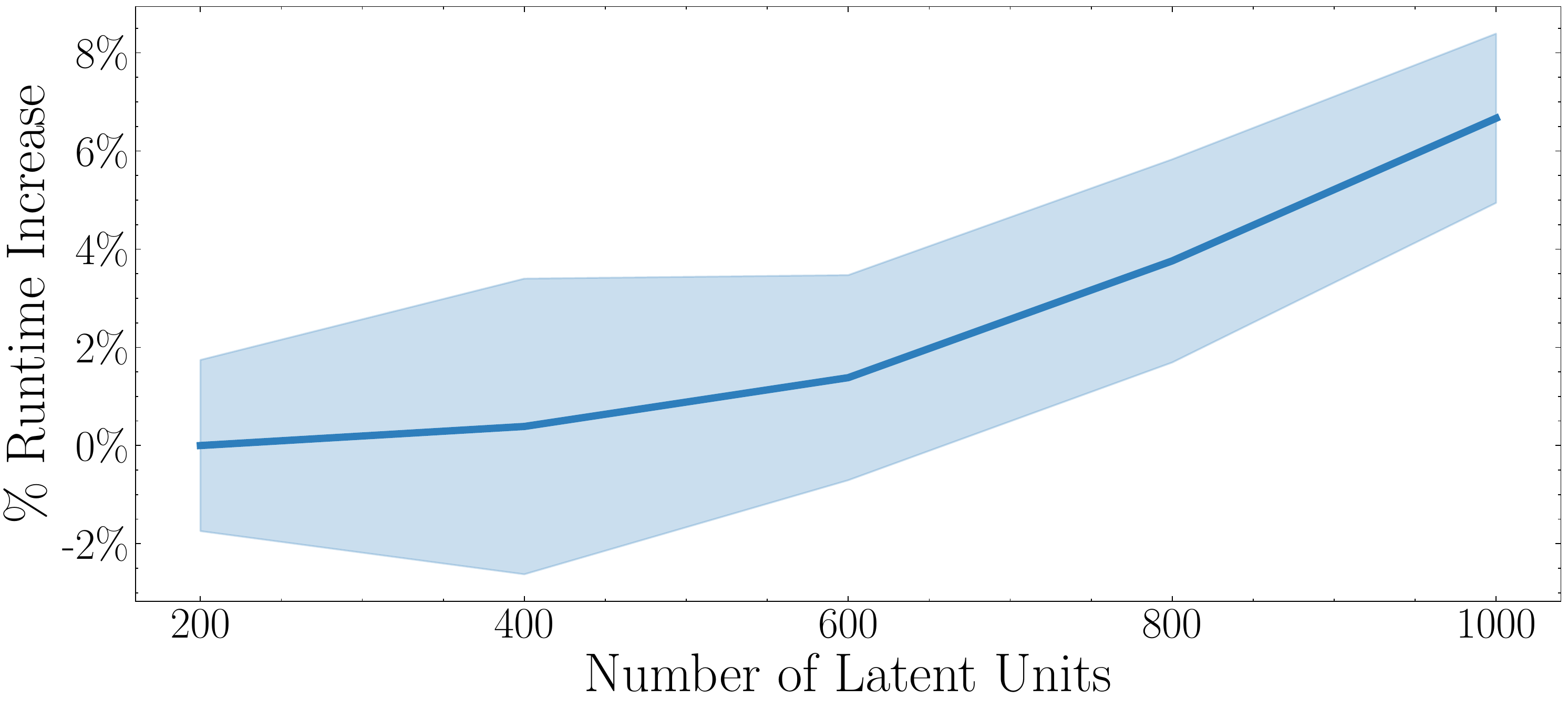}
  \caption{\textbf{Scaling to large models.} With a subsampling rate fixed at $M = 50$, \texttt{TANGOS} incurs only a small percentage increase in runtime as the number of neurons in the penultimate hidden layer increases dramatically.}
  \label{fig:scale}
  \vspace{-0.5em}
\end{figure}

Given the results in Figure \ref{fig:compute}, we next wish to evaluate if the proposed sampling scheme enables \texttt{TANGOS} to scale to much bigger models. In order to evaluate this we vary the number of neurons in the relevant hidden layer while maintaining a fixed sampling rate of 50 pairs (consistent with our experiments in Section \ref{sec:experiments}). Other experimental parameters are consistent with the previous experiment. The results are provided in Figure \ref{fig:scale} where we observe a relatively slow increase in runtime as the model grows. These results demonstrate that \texttt{TANGOS} can efficiently be applied to much larger models by using our proposed sampling scheme.

{\section{Ablation Study}} \label{appendix:ablation_study}

{\texttt{TANGOS} is designed with joint application of both regularization on specialization and orthogonalization in mind. Having empirically demonstrated strong overall results, an immediate question is the dynamics of the two regularizers, and how they interact to affect performance. Specifically, we consider the performance gain due to joint regularization effects over applying each regularizer separately.}

This includes three separate settings: $1)$ when the specialization regularizer is applied independently (\texttt{SpecOnly}), here we set $\lambda_2=0$ and search over $\lambda_1 \in \{1, 10, 100\}$; $2)$ when the orthogonalization is applied separately (\texttt{OrthOnly}), we set $\lambda_1=0$ and search over $\lambda_2\in\{0.1, 1\}$; and lastly $3)$ when both are applied jointly (\texttt{TANGOS}), i.e. searching over $\lambda_1\in \{1, 10, 100\}$ and $\lambda_2 \in \{0.1, 1\}$. We report the result of the ablation study in \Cref{tab:ablation_results}. We empirically observe that the joint effects of both regularizers (i.e. \texttt{TANGOS}) are crucial to achieve consistently good performance. 

Combining these results with what we observed in \Cref{fig:comparison_regularisation}, we hypothesize that applying just specialization regularization, with no regard for diversity, can inadvertently force the neurons to \emph{attend to overlapping regions} in the input space. Correspondingly, simply enforcing orthogonalization, with no regard for sparsity, will likely result in neurons attending to non-overlapping yet \emph{spurious} regions in the input. Thus, we conclude that the two regularizers have distinct, but complementary, effects that work together to achieve the desired regularization effect.

\begin{table}[h!]
\caption{{\textbf{Ablation study.} Generalization performance on different ablation settings.}}
\label{tab:ablation_results}
\centering
\midsepremove
\begin{tabular}{|c|ccc|ccc|}
\toprule
 & \multicolumn{3}{c|}{\cellcolor[HTML]{EFEFEF}Classification (Mean NLL)} & \multicolumn{3}{c|}{\cellcolor[HTML]{EFEFEF}Regression (MSE)} \\ \hline
\textbf{Dataset} & \textbf{BR} & \textbf{CR} & \textbf{HC} & \textbf{BC} & \textbf{BH} & \textbf{WQ} \\ \midrule
\texttt{NoReg} & $0.0726$ & $0.4633$ & $0.3321$ & $0.3343$ & $0.1977$ & $0.6732$ \\
\texttt{SpecOnly} & $0.0742$ & $0.4466$ & $0.3837$ & $0.3099$ & $0.1842$ & $0.6714$ \\
\texttt{OrthOnly} & $0.0716$ & $0.3696$ & $\mathbf{0.2073}$ & $0.2692$ & $0.1916$ & $0.6529$ \\
\texttt{TANGOS} & $\mathbf{0.070}$ & $\mathbf{0.3633}$ & $0.2191$ & $\mathbf{0.2472}$ & $\mathbf{0.1826}$ & $\mathbf{0.6379}$ \\
\bottomrule
\end{tabular}
\end{table}

\newpage

\section{Ranking Plot} \label{appdx:ranking_plot}
\begin{figure}[!htb]
  \includegraphics[width=\linewidth]{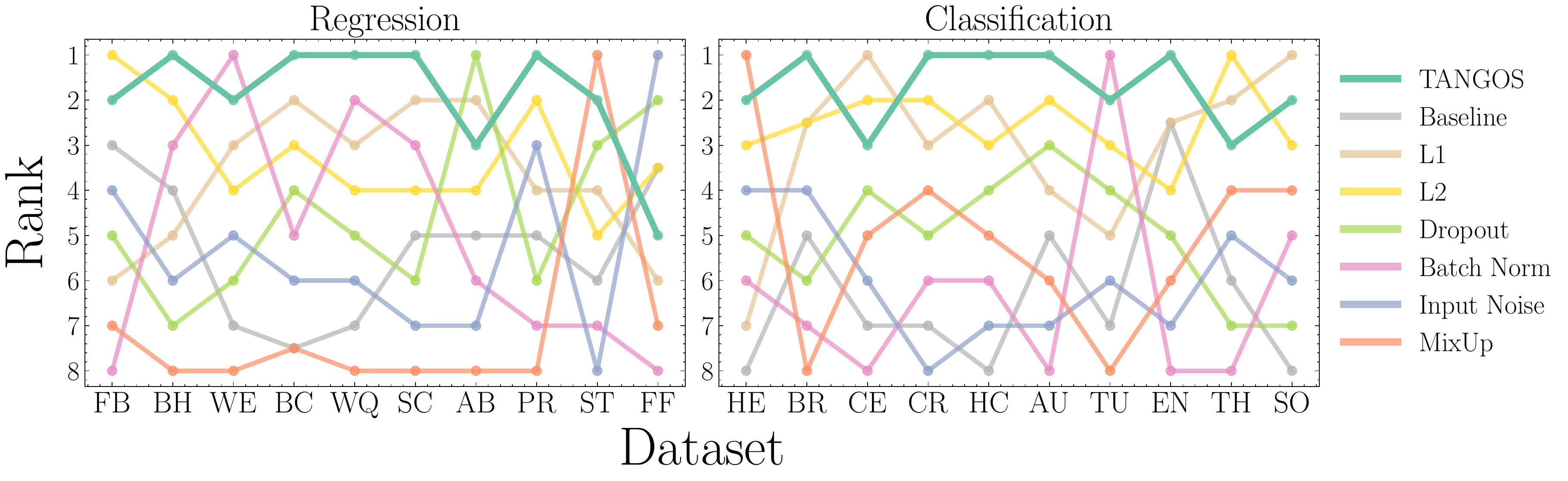}
  \caption{\textbf{Ranking of stand-alone regularizers.} Relative ranking of regularizer performance across the $20$ datasets, as reported in Table \ref{tab:stand_alone_generalisation}. \texttt{TANGOS} consistently ranks among the best-performing regularizers.}
  \label{fig:rank_plot}
  \vspace{-0.5em}
\end{figure}

In \Cref{tab:stand_alone_generalisation}, we reported the generalization performance of \texttt{TANGOS} compared to other regularizers in a stand-alone setting. To gain a better understanding of relative performance, we visually depict the relative ranking of regularizers across all $20$ datasets. \Cref{fig:rank_plot} demonstrates that \texttt{TANGOS} consistently ranks as one of the better-performing regularizers, while performance of benchmark methods tend to fluctuate depending on the dataset.

\section{Stand-Alone Uncertainty}
In \Cref{tab:stand_alone_uncertainty}, we report the standard deviation on generalization performance reported in \Cref{tab:stand_alone_generalisation}. The standard errors are computed using $10$ seeded runs.
\begin{table}[h!]
\vspace{-0.0cm}
\tiny
\centering
\caption{\textbf{Standard error on generalization performance.} Standard errors with respect to the random seed after retraining models from Table \ref{tab:stand_alone_generalisation} experiments $10$ times.}
\label{tab:stand_alone_uncertainty}
\resizebox{0.99\textwidth}{!}{%
\midsepremove
\begin{tabular}{|l|l|l|l|l|l|l|l|l|}
\toprule
Dataset & Baseline & L1 & L2 & DO & BN & IN & MU & TANGOS \\ \midrule 
\multicolumn{9}{c}{\cellcolor[HTML]{EFEFEF}Regression (Mean Squared Error)}                                                   \\ \midrule

FB &  0.051  &  0.016  &  0.009  &  0.051  &  0.627  &  0.076  &  0.041  &  0.042  \\
BH &  0.023  &  0.021  &  0.029  &  0.022  &  0.025  &  0.023  &  0.011  &  0.028  \\
WE &  0.006  &  0.010  &  0.008  &  0.008  &  0.008  &  0.013  &  0.010  &  0.009  \\
BC &  0.013  &  0.007  &  0.005  &  0.009  &  0.020  &  0.024  &  0.012  &  0.009  \\
WQ &  0.016  &  0.019  &  0.005  &  0.014  &  0.021  &  0.015  &  0.019  &  0.008  \\
SC &  0.026  &  0.014  &  0.019  &  0.025  &  0.023  &  0.168  &  0.067  &  0.017  \\
FF &  0.034  &  0.029  &  0.035  &  0.033  &  0.041  &  0.036  &  0.031  &  0.035  \\
PR &  0.042  &  0.029  &  0.020  &  0.031  &  0.032  &  0.072  &  0.016  &  0.026  \\
ST &  0.090  &  0.084  &  0.085  &  0.064  &  0.076  &  0.080  &  0.082  &  0.029  \\
AB &  0.016  &  0.016  &  0.008  &  0.011  &  0.026  &  0.014  &  0.012  &  0.006  \\

\midrule
\multicolumn{9}{c}{\cellcolor[HTML]{EFEFEF}Classification (Mean Negative Log-likelihood)}                                           \\ \midrule

HE &  0.057  &  0.049  &  0.009  &  0.033  &  0.163  &  0.038  &  0.067  &  0.032  \\
BR &  0.086  &  0.005  &  0.002  &  0.133  &  0.034  &  0.060  &  0.010  &  0.031  \\
CE &  0.060  &  0.007  &  0.008  &  0.043  &  0.051  &  0.044  &  0.056  &  0.023  \\
CR &  0.029  &  0.094  &  0.004  &  0.034  &  0.041  &  0.027  &  0.027  &  0.019  \\
HC &  0.091  &  0.014  &  0.019  &  0.026  &  0.054  &  0.111  &  0.024  &  0.014  \\
AU &  0.038  &  0.030  &  0.002  &  0.030  &  0.081  &  0.031  &  0.037  &  0.019  \\
TU &  0.075  &  0.075  &  0.087  &  0.077  &  0.087  &  0.096  &  0.078  &  0.064  \\
EN &  0.036  &  0.038  &  0.033  &  0.045  &  0.082  &  0.049  &  0.050  &  0.046  \\
TH &  0.048  &  0.021  &  0.002  &  0.065  &  0.096  &  0.047  &  0.039  &  0.001  \\
SO &  0.038  &  0.028  &  0.016  &  0.046  &  0.029  &  0.025  &  0.023  &  0.008  \\

\bottomrule
\end{tabular}%
\midsepdefault
}
\vspace{-0.0cm}
\end{table}

\section{Insights - Extended Results} \label{appdx:extended_results}
\begin{figure}[h!]
\centering
\begin{subfigure}{\textwidth}
\centering
  \includegraphics[width=0.75\linewidth]{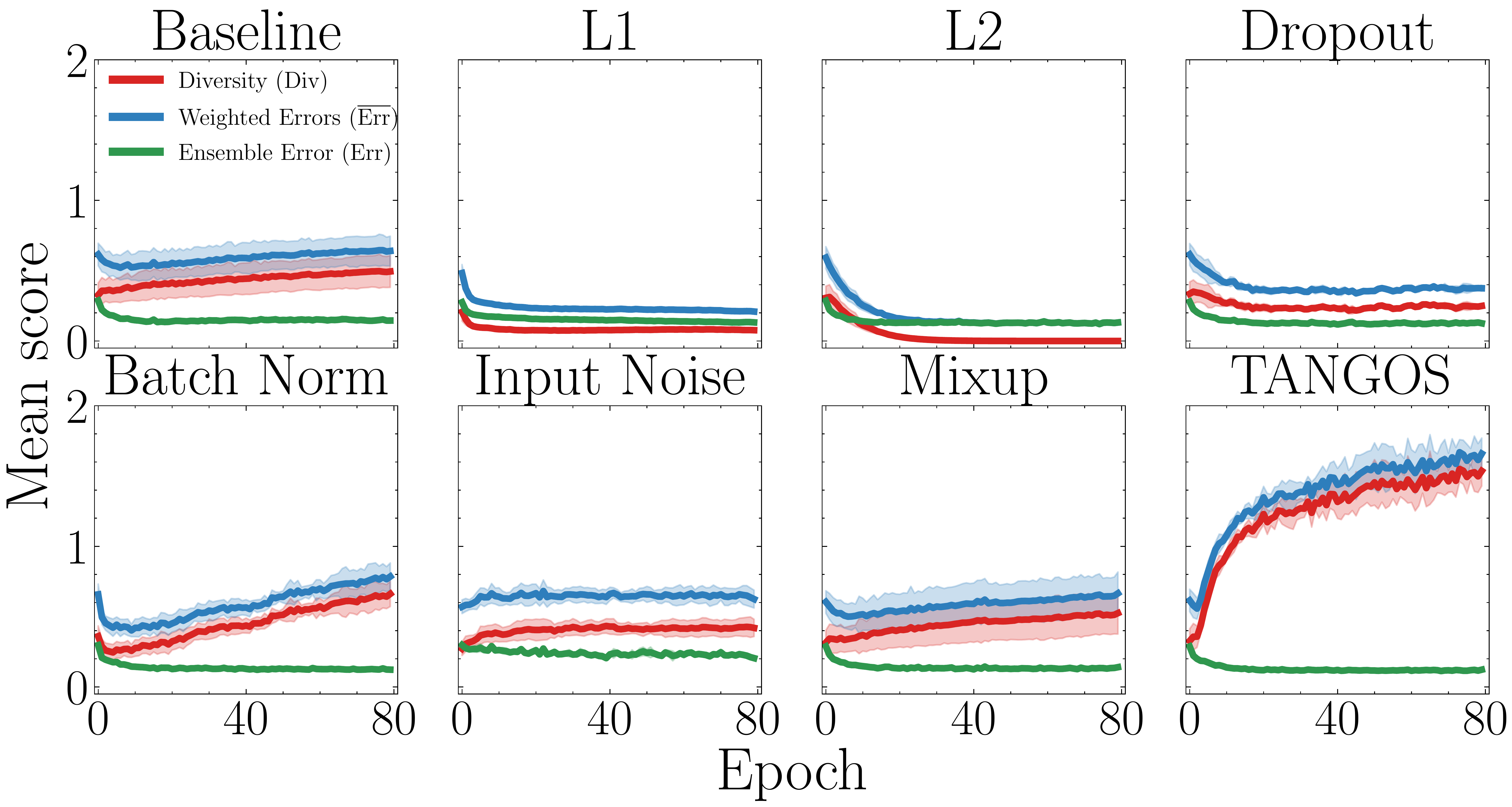}
\caption{Weather (WE) dataset.}
\end{subfigure}
\begin{subfigure}{\textwidth}
\centering
  \includegraphics[width=0.75\linewidth]{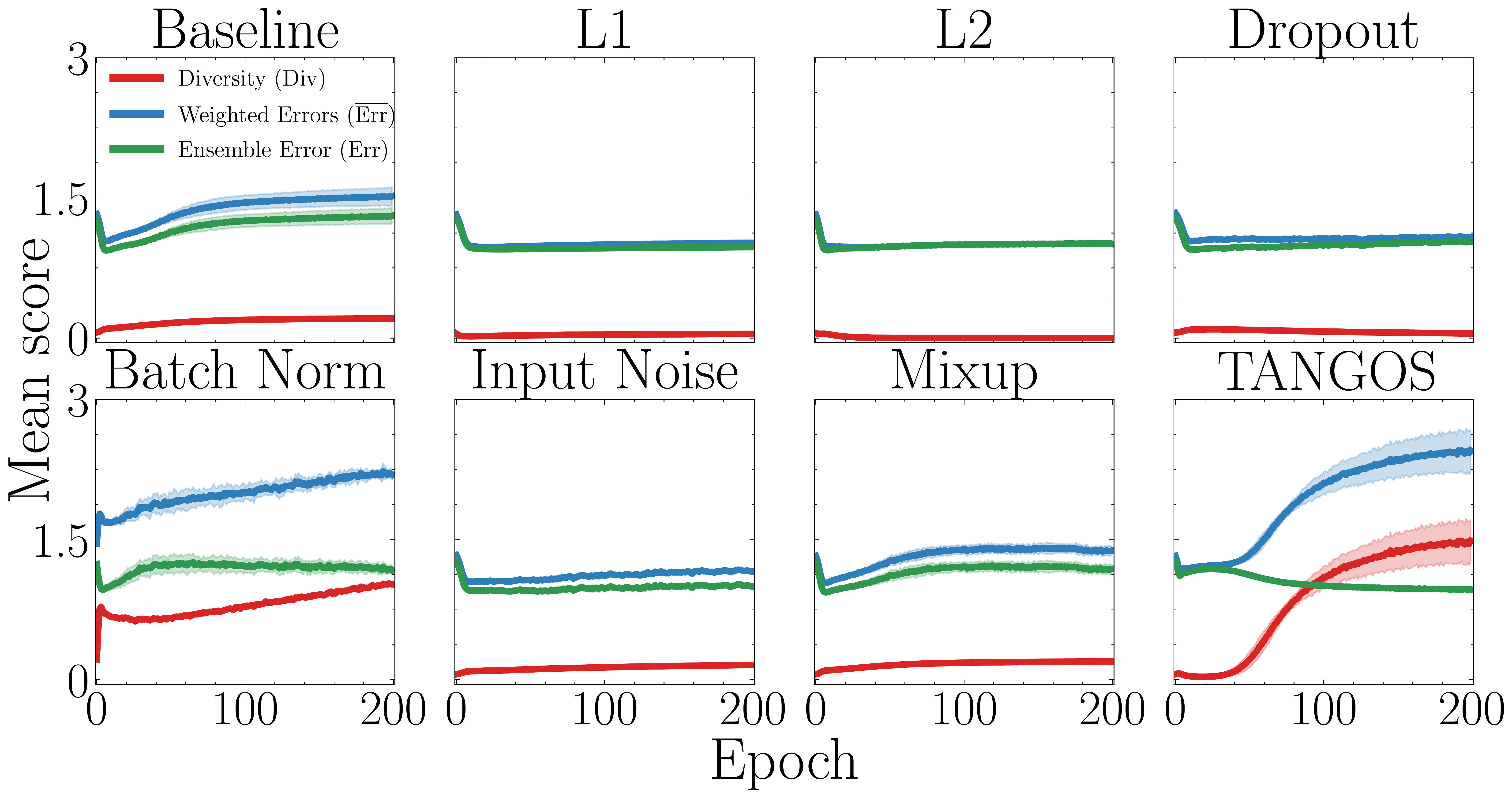}
\caption{Student (ST) dataset.}
\end{subfigure}
\begin{subfigure}{\textwidth}
\centering
  \includegraphics[width=0.75\linewidth]{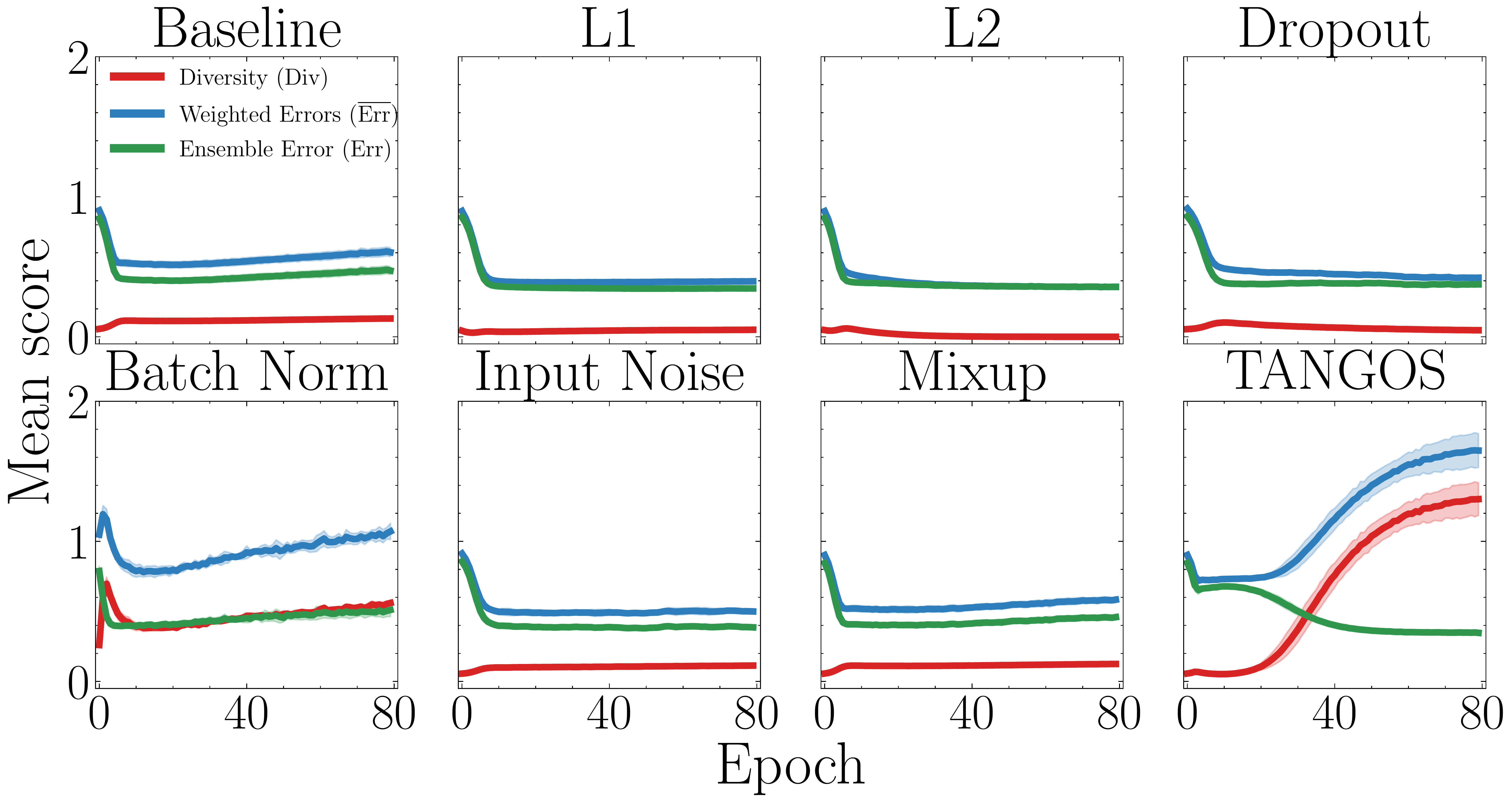}
\caption{Bioconcentration (BC) dataset.}
\end{subfigure}
\caption{\textbf{Neuron diversity.} Further examples of ensemble decomposition $\mathrm{Err} = \overline{\mathrm{Err}} - \mathrm{Div}$ as discussed in Section \ref{sec:diversity}.}
\label{fig:diversity_sup}
\end{figure}

In this section, we present extended results of the decomposition of overall model error into diversity and weighted error among an ensemble of latent units from Section \ref{sec:diversity}. We include all eight regularizers as described in Table \ref{tab:regularizers} and three datasets (WE, ST, and BC) as described in Table \ref{tab:data_summary}. The results are included in Figure \ref{fig:diversity_sup}.




\newpage

\section{In Tandem Results}
In \Cref{tab:paired_results}, we provide a detailed breakdown of \Cref{fig:in-tandem}, specifically by reporting in tandem performance when benchmarks are paired with \texttt{TANGOS} across all datasets. 
\begin{table}[h!]
\tiny
\centering
\caption{\textbf{In tandem performance.} Mean $\pm$ standard deviation of generalization performance when each regularizer is employed in tandem with \texttt{TANGOS}.}
\label{tab:paired_results}
\resizebox{\textwidth}{!}{%
\midsepremove
\begin{tabular}{|c|c|c|c|c|c|c|}
\toprule
Dataset & L1 & L2 & DO & BN & IN & MU \\ \midrule 
\multicolumn{7}{c}{\cellcolor[HTML]{EFEFEF}Regression (Mean Squared Error)}                                                   \\ \midrule
FB       &  0.033 $\pm{0.018}$  &  0.018 $\pm{0.009}$  &  0.023 $\pm{0.215}$  &  0.042 $\pm{0.268}$  &  0.028 $\pm{0.047}$  &  0.046 $\pm{0.073}$  \\
BH       &  0.192 $\pm{0.022}$  &  0.176 $\pm{0.024}$  &  0.196 $\pm{0.024}$  &  0.220 $\pm{0.021}$  &  0.191 $\pm{0.023}$  &  0.178 $\pm{0.019}$  \\
WE       &  0.093 $\pm{0.011}$  &  0.091 $\pm{0.009}$  &  0.092 $\pm{0.009}$  &  0.076 $\pm{0.009}$  &  0.081 $\pm{0.012}$  &  0.077 $\pm{0.013}$  \\
WQ       &  0.637 $\pm{0.014}$  &  0.639 $\pm{0.006}$  &  0.628 $\pm{0.008}$  &  0.630 $\pm{0.015}$  &  0.644 $\pm{0.011}$  &  0.649 $\pm{0.018}$  \\
BC       &  0.227 $\pm{0.011}$  &  0.236 $\pm{0.011}$  &  0.243 $\pm{0.013}$  &  0.275 $\pm{0.027}$  &  0.235 $\pm{0.009}$  &  0.260 $\pm{0.014}$  \\
SC       &  0.407 $\pm{0.044}$  &  0.399 $\pm{0.072}$  &  0.370 $\pm{0.031}$  &  0.408 $\pm{0.049}$  &  0.391 $\pm{0.317}$  &  0.425 $\pm{0.061}$  \\
AB       &  0.309 $\pm{0.028}$  &  0.312 $\pm{0.028}$  &  0.312 $\pm{0.026}$  &  0.311 $\pm{0.027}$  &  0.319 $\pm{0.037}$  &  0.308 $\pm{0.03}$  \\
FF       &  1.281 $\pm{0.029}$  &  1.276 $\pm{0.043}$  &  1.268 $\pm{0.028}$  &  1.297 $\pm{0.027}$  &  1.203 $\pm{0.046}$  &  1.207 $\pm{0.011}$  \\
PR       &  0.553 $\pm{0.047}$  &  0.572 $\pm{0.031}$  &  0.642 $\pm{0.093}$  &  0.561 $\pm{0.069}$  &  0.565 $\pm{0.025}$  &  0.568 $\pm{0.081}$  \\
ST       &  0.392 $\pm{0.006}$  &  0.382 $\pm{0.006}$  &  0.388 $\pm{0.016}$  &  0.446 $\pm{0.018}$  &  0.417 $\pm{0.012}$  &  0.447 $\pm{0.011}$  \\

\midrule
\multicolumn{7}{c}{\cellcolor[HTML]{EFEFEF}Classification (Mean Negative Log-likelihood)}                                           \\ \midrule

HE      &  0.441 $\pm{0.046}$  &  0.427 $\pm{0.049}$  &  0.454 $\pm{0.095}$  &  0.407 $\pm{0.047}$  &  0.377 $\pm{0.067}$  &  0.397 $\pm{0.075}$  \\
BR      &  0.074 $\pm{0.005}$  &  0.070 $\pm{0.002}$  &  0.068 $\pm{0.006}$  &  0.062 $\pm{0.003}$  &  0.065 $\pm{0.01}$  &  0.078 $\pm{0.011}$  \\
CE      &  0.389 $\pm{0.007}$  &  0.396 $\pm{0.007}$  &  0.394 $\pm{0.033}$  &  0.446 $\pm{0.024}$  &  0.422 $\pm{0.075}$  &  0.422 $\pm{0.062}$  \\
CR      &  0.362 $\pm{0.021}$  &  0.366 $\pm{0.015}$  &  0.364 $\pm{0.003}$  &  0.406 $\pm{0.023}$  &  0.367 $\pm{0.01}$  &  0.384 $\pm{0.036}$  \\
HC      &  0.200 $\pm{0.056}$  &  0.179 $\pm{0.012}$  &  0.185 $\pm{0.009}$  &  0.186 $\pm{0.021}$  &  0.211 $\pm{0.008}$  &  0.181 $\pm{0.037}$  \\
AU      &  0.368 $\pm{0.028}$  &  0.360 $\pm{0.138}$  &  0.352 $\pm{0.011}$  &  0.344 $\pm{0.013}$  &  0.379 $\pm{0.016}$  &  0.368 $\pm{0.041}$  \\
TU      &  1.519 $\pm{0.082}$  &  1.506 $\pm{0.045}$  &  1.481 $\pm{0.048}$  &  1.506 $\pm{0.067}$  &  1.522 $\pm{0.049}$  &  1.503 $\pm{0.087}$  \\
SO      &  0.227 $\pm{0.036}$  &  0.268 $\pm{0.071}$  &  0.233 $\pm{0.021}$  &  0.353 $\pm{0.052}$  &  0.257 $\pm{0.032}$  &  0.304 $\pm{0.054}$  \\
EN      &  0.990 $\pm{0.024}$  &  0.971 $\pm{0.002}$  &  0.945 $\pm{0.001}$  &  1.004 $\pm{0.026}$  &  0.995 $\pm{0.001}$  &  1.007 $\pm{0.04}$  \\
TH      &  0.506 $\pm{0.031}$  &  0.503 $\pm{0.008}$  &  0.513 $\pm{0.01}$  &  0.524 $\pm{0.008}$  &  0.512 $\pm{0.045}$  &  0.514 $\pm{0.03}$  \\
\bottomrule
\end{tabular}%
\midsepdefault
}
\end{table}

\section{Dataset and Regularizer Details} \label{appndx:data}
We perform our experiments on 20 real-world publicly available datasets obtained from \citep{dua2017uci}. They are summarized in Table \ref{tab:data_summary}. 
Further information and the source files used for each of the respective datasets can be found at: \url{https://archive.ics.uci.edu/ml/machine-learning-databases/<UCI Source>/} where \texttt{<UCI Source>} denotes the datasets unique identifier as listed in Table \ref{tab:data_summary}. Standard preprocessing was applied including standardization of features, one hot encoding of categorical variables, median imputation of missing values and log transformations of highly skewed feature distributions. Furthermore, for computational feasibility, datasets with over $1000$ samples were reduced in size. In these cases the first $1000$ samples from the original UCI Source file were used. In Table \ref{tab:regularizers} we summarize the regularizers considered in this work.

\begin{table}[h!]
\centering
\caption{\textbf{Dataset descriptions.} Summary of the datasets considered in this work.}
\label{tab:data_summary}
\resizebox{0.99\textwidth}{!}{%
\begin{tabular}{c c c c c c } 
\hline
\multicolumn{1}{c}{Dataset} & \multicolumn{1}{c}{UCI Source} & \multicolumn{1}{c}{Type} & \multicolumn{1}{c}{Feature size} & \multicolumn{1}{c}{Sample Size} & \multicolumn{1}{c}{Reference} \\ \hline
Facebook (FB) & ``00368'' & Regression & 21 & 495 & \cite{moro2016predicting} \\
Boston (BH) & [1] & Regression & 13 & 506 & \cite{harrison1978hedonic} \\
Weather (WE) & ``00514'' & Regression & 45 & 1000 & \cite{cho2020comparative} \\
Wine Quality (WQ) & ``wine-quality'' & Regression & 11 & 1000 & \cite{cortez2009modeling} \\
Bioconcentration (BC) & ``00510'' & Regression & 45 & 779 & \cite{grisoni2015qsar} \\
Skillcraft (SC) & ``00272'' & Regression & 18 & 1000 & \cite{thompson2013video} \\
Forest Fire (FF) & ``forest-fires'' & Regression & 39 & 517 & \cite{cortez2007data} \\
Protein (PR) & ``00265'' & Regression & 9 & 1000 & \cite{dua2017uci} \\
Student (ST) & ``00320'' & Regression & 56 & 649 & \cite{cortez2008using} \\
Abalone (AB) & ``abalone'' & Regression & 9 & 1000 & \cite{waugh1995extending} \\
Heart (HE) & ``statlog'' & Classification & 20 & 270 & \cite{dua2017uci} \\
Breast (BR) & ``breast-cancer-wisconsin'' & Classification & 9 & 699 & \cite{street1993nuclear} \\
Cervical (CE) & ``00383'' & Classification & 136 & 858 & \cite{fernandes2017transfer} \\
Credit (CR) & ``credit-screening'' & Classification & 40 & 677 & \cite{dua2017uci} \\
HCV (HC) & ``00571'' & Classification & 12 & 615 & \cite{hoffmann2018using} \\
Australian (AU) & ``statlog'' & Classification & 55 & 690 & \cite{quinlan1987simplifying} \\
Tumor (TU) & ``primary-tumor'' & Classification & 25 & 339 & \cite{michalski1986multi} \\
Entrance (EN) & ``00582'' & Classification & 38 & 666 & \cite{hussain2018classification} \\
Thoracic (TH) & ``00277'' & Classification & 24 & 470 & \cite{zieba2013boosted} \\
Soybean (SO) & ``soybean'' & Classification & 484 & 683 & \cite{fisher1988concept} \\

\hline
\end{tabular}}
\label{default}
\begin{tablenotes}
\footnotesize
\item[] [1] This dataset has now been removed due to ethical issues. For more information see the following url \url{https://medium.com/@docintangible/racist-data-destruction-113e3eff54a8} 
\end{tablenotes}
\end{table}

\begin{table}[h!]
\caption{\textbf{Overview of regularizers.} Description of benchmarks considered in this work and their implementations.}
\centering
\begin{tabular}{ccc}
\hline
Regularizer & Reference & Implementation \\
\hline
Baseline & NA & \cite{paszke2019pytorch} \\ 
L1 & \cite{tibshirani1996regression} & \cite{paszke2019pytorch} \\
L2 & \cite{hoerl1970ridge} & \cite{paszke2019pytorch} \\
Dropout & \cite{hinton2012improving} & \cite{paszke2019pytorch} \\
Batch Norm & \cite{ioffe2015batch} & \cite{paszke2019pytorch} \\ 
Input Noise & \cite{krizhevsky2012imagenet} & \cite{paszke2019pytorch} \\ 
Mixup & \cite{zhang2017mixup} & \cite{zhang2017mixup} \\ 
TANGOS & This work & \cite{paszke2019pytorch} \\ 
\hline
\end{tabular}
\label{tab:regularizers}
\end{table}

\end{document}

%% file: tex_files/math_commands.tex
\usepackage{amsmath,amsfonts,bm}

\def\eqref#1{equation~\ref{#1}}

\def\1{\bm{1}}

\DeclareMathAlphabet{\mathsfit}{\encodingdefault}{\sfdefault}{m}{sl}
\SetMathAlphabet{\mathsfit}{bold}{\encodingdefault}{\sfdefault}{bx}{n}

\DeclareMathOperator*{\argmin}{arg\,min}

%% file: camera_ready.bbl
\begin{thebibliography}{89}
\providecommand{\natexlab}[1]{#1}
\providecommand{\url}[1]{\texttt{#1}}
\expandafter\ifx\csname urlstyle\endcsname\relax
  \providecommand{\doi}[1]{doi: #1}\else
  \providecommand{\doi}{doi: \begingroup \urlstyle{rm}\Url}\fi

\bibitem[Acosta et~al.(2022)Acosta, Falcone, Rajpurkar, and
  Topol]{acosta2022multimodal}
Juli{\'a}n~N Acosta, Guido~J Falcone, Pranav Rajpurkar, and Eric~J Topol.
\newblock Multimodal biomedical {AI}.
\newblock \emph{Nature Medicine}, 28\penalty0 (9):\penalty0 1773--1784, 2022.

\bibitem[Ancona et~al.(2017)Ancona, Ceolini, {\"O}ztireli, and
  Gross]{ancona2017towards}
Marco Ancona, Enea Ceolini, Cengiz {\"O}ztireli, and Markus Gross.
\newblock Towards better understanding of gradient-based attribution methods
  for deep neural networks.
\newblock \emph{arXiv preprint arXiv:1711.06104}, 2017.

\bibitem[Bahri et~al.(2021)Bahri, Jiang, Tay, and Metzler]{bahri2021scarf}
Dara Bahri, Heinrich Jiang, Yi~Tay, and Donald Metzler.
\newblock {SCARF}: Self-supervised contrastive learning using random feature
  corruption.
\newblock \emph{arXiv preprint arXiv:2106.15147}, 2021.

\bibitem[Baldi et~al.(2014)Baldi, Sadowski, and Whiteson]{baldi2014searching}
Pierre Baldi, Peter Sadowski, and Daniel Whiteson.
\newblock Searching for exotic particles in high-energy physics with deep
  learning.
\newblock \emph{Nature Communications}, 5\penalty0 (1):\penalty0 1--9, 2014.

\bibitem[Bansal et~al.(2018)Bansal, Chen, and Wang]{bansal2018can}
Nitin Bansal, Xiaohan Chen, and Zhangyang Wang.
\newblock Can we gain more from orthogonality regularizations in training deep
  networks?
\newblock \emph{Advances in Neural Information Processing Systems}, 31, 2018.

\bibitem[Breiman(1996)]{breiman1996bagging}
Leo Breiman.
\newblock Bagging predictors.
\newblock \emph{Machine learning}, 24\penalty0 (2):\penalty0 123--140, 1996.

\bibitem[Breiman(2001)]{breiman2001random}
Leo Breiman.
\newblock Random forests.
\newblock \emph{Machine learning}, 45\penalty0 (1):\penalty0 5--32, 2001.

\bibitem[Brigato \& Iocchi(2021)Brigato and Iocchi]{brigato2021close}
Lorenzo Brigato and Luca Iocchi.
\newblock A close look at deep learning with small data.
\newblock In \emph{2020 25th International Conference on Pattern Recognition
  (ICPR)}, pp.\  2490--2497. IEEE, 2021.

\bibitem[B{\"u}hlmann(2012)]{buhlmann2012bagging}
Peter B{\"u}hlmann.
\newblock Bagging, boosting and ensemble methods.
\newblock In \emph{Handbook of computational statistics}, pp.\  985--1022.
  Springer, 2012.

\bibitem[Chen et~al.(2019)Chen, Wu, Rastogi, Liang, and Jha]{chen2019robust}
Jiefeng Chen, Xi~Wu, Vaibhav Rastogi, Yingyu Liang, and Somesh Jha.
\newblock Robust attribution regularization.
\newblock \emph{Advances in Neural Information Processing Systems}, 32, 2019.

\bibitem[Chen \& Guestrin(2016)Chen and Guestrin]{chen2016xgboost}
Tianqi Chen and Carlos Guestrin.
\newblock Xgboost: A scalable tree boosting system.
\newblock In \emph{Proceedings of the 22nd acm sigkdd international conference
  on knowledge discovery and data mining}, pp.\  785--794, 2016.

\bibitem[Cho et~al.(2020)Cho, Yoo, Im, and Cha]{cho2020comparative}
Dongjin Cho, Cheolhee Yoo, Jungho Im, and Dong-Hyun Cha.
\newblock Comparative assessment of various machine learning-based bias
  correction methods for numerical weather prediction model forecasts of
  extreme air temperatures in urban areas.
\newblock \emph{Earth and Space Science}, 7\penalty0 (4):\penalty0
  e2019EA000740, 2020.

\bibitem[Cortez \& Morais(2007)Cortez and Morais]{cortez2007data}
Paulo Cortez and An{\'\i}bal de Jesus~Raimundo Morais.
\newblock A data mining approach to predict forest fires using meteorological
  data.
\newblock 2007.

\bibitem[Cortez \& Silva(2008)Cortez and Silva]{cortez2008using}
Paulo Cortez and Alice Maria~Gon{\c{c}}alves Silva.
\newblock Using data mining to predict secondary school student performance.
\newblock 2008.

\bibitem[Cortez et~al.(2009)Cortez, Cerdeira, Almeida, Matos, and
  Reis]{cortez2009modeling}
Paulo Cortez, Ant{\'o}nio Cerdeira, Fernando Almeida, Telmo Matos, and Jos{\'e}
  Reis.
\newblock Modeling wine preferences by data mining from physicochemical
  properties.
\newblock \emph{Decision support systems}, 47\penalty0 (4):\penalty0 547--553,
  2009.

\bibitem[Cowan et~al.(2005)Cowan, Elliott, Saults, Morey, Mattox,
  Hismjatullina, and Conway]{cowan2005capacity}
Nelson Cowan, Emily~M Elliott, J~Scott Saults, Candice~C Morey, Sam Mattox,
  Anna Hismjatullina, and Andrew~RA Conway.
\newblock On the capacity of attention: Its estimation and its role in working
  memory and cognitive aptitudes.
\newblock \emph{Cognitive psychology}, 51\penalty0 (1):\penalty0 42--100, 2005.

\bibitem[Crabb{\'e} \& van~der Schaar(2022)Crabb{\'e} and van~der
  Schaar]{crabbe2022label}
Jonathan Crabb{\'e} and Mihaela van~der Schaar.
\newblock Label-free explainability for unsupervised models.
\newblock \emph{arXiv preprint arXiv:2203.01928}, 2022.

\bibitem[Crabb{\'e} et~al.(2021)Crabb{\'e}, Qian, Imrie, and van~der
  Schaar]{crabbe2021explaining}
Jonathan Crabb{\'e}, Zhaozhi Qian, Fergus Imrie, and Mihaela van~der Schaar.
\newblock Explaining latent representations with a corpus of examples.
\newblock \emph{Advances in Neural Information Processing Systems},
  34:\penalty0 12154--12166, 2021.

\bibitem[Drucker \& Le~Cun(1992)Drucker and Le~Cun]{drucker1992improving}
Harris Drucker and Yann Le~Cun.
\newblock Improving generalization performance using double backpropagation.
\newblock \emph{IEEE Transactions on Neural Networks}, 3\penalty0 (6):\penalty0
  991--997, 1992.

\bibitem[Dua et~al.(2017)Dua, Graff, et~al.]{dua2017uci}
Dheeru Dua, Casey Graff, et~al.
\newblock {UCI} machine learning repository.
\newblock 2017.

\bibitem[Erion et~al.(2021)Erion, Janizek, Sturmfels, Lundberg, and
  Lee]{erion2021improving}
Gabriel Erion, Joseph~D Janizek, Pascal Sturmfels, Scott~M Lundberg, and Su-In
  Lee.
\newblock Improving performance of deep learning models with axiomatic
  attribution priors and expected gradients.
\newblock \emph{Nature machine intelligence}, 3\penalty0 (7):\penalty0
  620--631, 2021.

\bibitem[Fernandes et~al.(2017)Fernandes, Cardoso, and
  Fernandes]{fernandes2017transfer}
Kelwin Fernandes, Jaime~S Cardoso, and Jessica Fernandes.
\newblock Transfer learning with partial observability applied to cervical
  cancer screening.
\newblock In \emph{Iberian conference on pattern recognition and image
  analysis}, pp.\  243--250. Springer, 2017.

\bibitem[Fisher \& Schlimmer(1988)Fisher and Schlimmer]{fisher1988concept}
Douglas~H Fisher and Jeffrey~C Schlimmer.
\newblock Concept simplification and prediction accuracy.
\newblock In \emph{Machine Learning Proceedings 1988}, pp.\  22--28. Elsevier,
  1988.

\bibitem[Gorishniy et~al.(2022)Gorishniy, Rubachev, and
  Babenko]{gorishniy2022embeddings}
Yura Gorishniy, Ivan Rubachev, and Artem Babenko.
\newblock On embeddings for numerical features in tabular deep learning.
\newblock \emph{arXiv preprint arXiv:2203.05556}, 2022.

\bibitem[Gorishniy et~al.(2021)Gorishniy, Rubachev, Khrulkov, and
  Babenko]{gorishniy2021revisiting}
Yury Gorishniy, Ivan Rubachev, Valentin Khrulkov, and Artem Babenko.
\newblock Revisiting deep learning models for tabular data.
\newblock \emph{Advances in Neural Information Processing Systems},
  34:\penalty0 18932--18943, 2021.

\bibitem[Grinsztajn et~al.(2022)Grinsztajn, Oyallon, and
  Varoquaux]{grinsztajn2022tree}
L{\'e}o Grinsztajn, Edouard Oyallon, and Ga{\"e}l Varoquaux.
\newblock Why do tree-based models still outperform deep learning on tabular
  data?
\newblock \emph{arXiv preprint arXiv:2207.08815}, 2022.

\bibitem[Grisoni et~al.(2015)Grisoni, Consonni, Villa, Vighi, and
  Todeschini]{grisoni2015qsar}
Francesca Grisoni, Viviana Consonni, Sara Villa, Marco Vighi, and Roberto
  Todeschini.
\newblock Qsar models for bioconcentration: is the increase in the complexity
  justified by more accurate predictions?
\newblock \emph{Chemosphere}, 127:\penalty0 171--179, 2015.

\bibitem[Gulrajani et~al.(2017)Gulrajani, Ahmed, Arjovsky, Dumoulin, and
  Courville]{gulrajani2017improved}
Ishaan Gulrajani, Faruk Ahmed, Martin Arjovsky, Vincent Dumoulin, and Aaron~C
  Courville.
\newblock Improved training of {W}asserstein {GANs}.
\newblock \emph{Advances in Neural Information Processing Systems}, 30, 2017.

\bibitem[Guo et~al.(2019)Guo, Wang, and Wang]{guo2019deep}
Wenzhong Guo, Jianwen Wang, and Shiping Wang.
\newblock Deep multimodal representation learning: A survey.
\newblock \emph{IEEE Access}, 7:\penalty0 63373--63394, 2019.

\bibitem[Guyon et~al.(2019)Guyon, Sun-Hosoya, Boull{\'e}, Escalante, Escalera,
  Liu, Jajetic, Ray, Saeed, Sebag, et~al.]{guyon2019analysis}
Isabelle Guyon, Lisheng Sun-Hosoya, Marc Boull{\'e}, Hugo~Jair Escalante,
  Sergio Escalera, Zhengying Liu, Damir Jajetic, Bisakha Ray, Mehreen Saeed,
  Mich{\`e}le Sebag, et~al.
\newblock Analysis of the automl challenge series.
\newblock \emph{Automated Machine Learning}, pp.\  177, 2019.

\bibitem[Harrison~Jr \& Rubinfeld(1978)Harrison~Jr and
  Rubinfeld]{harrison1978hedonic}
David Harrison~Jr and Daniel~L Rubinfeld.
\newblock Hedonic housing prices and the demand for clean air.
\newblock \emph{Journal of environmental economics and management}, 5\penalty0
  (1):\penalty0 81--102, 1978.

\bibitem[He et~al.(2015)He, Zhang, Ren, and Sun]{he2015delving}
Kaiming He, Xiangyu Zhang, Shaoqing Ren, and Jian Sun.
\newblock Delving deep into rectifiers: Surpassing human-level performance on
  imagenet classification.
\newblock In \emph{Proceedings of the IEEE international conference on computer
  vision}, pp.\  1026--1034, 2015.

\bibitem[He et~al.(2022)He, Kang, Luo, Fan, and Yang]{he2022hybrid}
Yuanqin He, Yan Kang, Jiahuan Luo, Lixin Fan, and Qiang Yang.
\newblock A hybrid self-supervised learning framework for vertical federated
  learning.
\newblock \emph{arXiv preprint arXiv:2208.08934}, 2022.

\bibitem[Hinton et~al.(2012)Hinton, Srivastava, Krizhevsky, Sutskever, and
  Salakhutdinov]{hinton2012improving}
Geoffrey~E Hinton, Nitish Srivastava, Alex Krizhevsky, Ilya Sutskever, and
  Ruslan~R Salakhutdinov.
\newblock Improving neural networks by preventing co-adaptation of feature
  detectors.
\newblock \emph{arXiv preprint arXiv:1207.0580}, 2012.

\bibitem[Hoerl \& Kennard(1970)Hoerl and Kennard]{hoerl1970ridge}
Arthur~E Hoerl and Robert~W Kennard.
\newblock Ridge regression: Biased estimation for nonorthogonal problems.
\newblock \emph{Technometrics}, 12\penalty0 (1):\penalty0 55--67, 1970.

\bibitem[Hoffmann et~al.(2018)Hoffmann, Bietenbeck, Lichtinghagen, and
  Klawonn]{hoffmann2018using}
Georg Hoffmann, Andreas Bietenbeck, Ralf Lichtinghagen, and Frank Klawonn.
\newblock Using machine learning techniques to generate laboratory diagnostic
  pathways—a case study.
\newblock \emph{J Lab Precis Med}, 3:\penalty0 58, 2018.

\bibitem[Hospedales et~al.(2021)Hospedales, Antoniou, Micaelli, and
  Storkey]{hospedales2021meta}
Timothy Hospedales, Antreas Antoniou, Paul Micaelli, and Amos Storkey.
\newblock Meta-learning in neural networks: A survey.
\newblock \emph{IEEE transactions on pattern analysis and machine
  intelligence}, 44\penalty0 (9):\penalty0 5149--5169, 2021.

\bibitem[Hu et~al.(2017)Hu, Peng, Yang, Hospedales, and
  Verbeek]{hu2017frankenstein}
Guosheng Hu, Xiaojiang Peng, Yongxin Yang, Timothy~M Hospedales, and Jakob
  Verbeek.
\newblock Frankenstein: Learning deep face representations using small data.
\newblock \emph{IEEE Transactions on Image Processing}, 27\penalty0
  (1):\penalty0 293--303, 2017.

\bibitem[Hussain et~al.(2018)Hussain, Atallah, Kamsin, and
  Hazarika]{hussain2018classification}
Sadiq Hussain, Rasha Atallah, Amirrudin Kamsin, and Jiten Hazarika.
\newblock Classification, clustering and association rule mining in educational
  datasets using data mining tools: A case study.
\newblock In \emph{Computer Science On-line Conference}, pp.\  196--211.
  Springer, 2018.

\bibitem[Ioffe \& Szegedy(2015)Ioffe and Szegedy]{ioffe2015batch}
Sergey Ioffe and Christian Szegedy.
\newblock Batch normalization: Accelerating deep network training by reducing
  internal covariate shift.
\newblock In \emph{International conference on machine learning}, pp.\
  448--456. PMLR, 2015.

\bibitem[Jin et~al.(2020)Jin, Yi, Zhang, Zhang, Schewe, and Huang]{jin2020does}
Gaojie Jin, Xinping Yi, Liang Zhang, Lijun Zhang, Sven Schewe, and Xiaowei
  Huang.
\newblock How does weight correlation affect generalisation ability of deep
  neural networks?
\newblock \emph{Advances in Neural Information Processing Systems},
  33:\penalty0 21346--21356, 2020.

\bibitem[Johnston \& Dark(1986)Johnston and Dark]{johnston1986selective}
William~A Johnston and Veronica~J Dark.
\newblock Selective attention.
\newblock \emph{Annual review of psychology}, 37\penalty0 (1):\penalty0 43--75,
  1986.

\bibitem[Kadra et~al.(2021)Kadra, Lindauer, Hutter, and
  Grabocka]{kadra2021well}
Arlind Kadra, Marius Lindauer, Frank Hutter, and Josif Grabocka.
\newblock Well-tuned simple nets excel on tabular datasets.
\newblock \emph{Advances in Neural Information Processing Systems}, 34, 2021.

\bibitem[Kim et~al.(2018)Kim, Wattenberg, Gilmer, Cai, Wexler, Viegas,
  et~al.]{kim2018interpretability}
Been Kim, Martin Wattenberg, Justin Gilmer, Carrie Cai, James Wexler, Fernanda
  Viegas, et~al.
\newblock Interpretability beyond feature attribution: Quantitative testing
  with concept activation vectors (tcav).
\newblock In \emph{International conference on machine learning}, pp.\
  2668--2677. PMLR, 2018.

\bibitem[Krizhevsky et~al.(2012)Krizhevsky, Sutskever, and
  Hinton]{krizhevsky2012imagenet}
Alex Krizhevsky, Ilya Sutskever, and Geoffrey~E Hinton.
\newblock Imagenet classification with deep convolutional neural networks.
\newblock \emph{Advances in Neural Information Processing Systems}, 25, 2012.

\bibitem[Krogh \& Vedelsby(1994)Krogh and Vedelsby]{krogh1994neural}
Anders Krogh and Jesper Vedelsby.
\newblock Neural network ensembles, cross validation, and active learning.
\newblock \emph{Advances in Neural Information Processing Systems}, 7, 1994.

\bibitem[Kuka{\v{c}}ka et~al.(2017)Kuka{\v{c}}ka, Golkov, and
  Cremers]{kukavcka2017regularization}
Jan Kuka{\v{c}}ka, Vladimir Golkov, and Daniel Cremers.
\newblock Regularization for deep learning: A taxonomy.
\newblock \emph{arXiv preprint arXiv:1710.10686}, 2017.

\bibitem[LeCun et~al.(1998)LeCun, Bottou, Bengio, and
  Haffner]{lecun1998gradient}
Yann LeCun, L{\'e}on Bottou, Yoshua Bengio, and Patrick Haffner.
\newblock Gradient-based learning applied to document recognition.
\newblock \emph{Proceedings of the IEEE}, 86\penalty0 (11):\penalty0
  2278--2324, 1998.

\bibitem[Lee et~al.(2022)Lee, Imrie, and van~der Schaar]{lee2022self}
Changhee Lee, Fergus Imrie, and Mihaela van~der Schaar.
\newblock Self-supervision enhanced feature selection with correlated gates.
\newblock In \emph{International Conference on Learning Representations}, 2022.

\bibitem[Levin et~al.(2022)Levin, Cherepanova, Schwarzschild, Bansal, Bruss,
  Goldstein, Wilson, and Goldblum]{levin2022transfer}
Roman Levin, Valeriia Cherepanova, Avi Schwarzschild, Arpit Bansal, C~Bayan
  Bruss, Tom Goldstein, Andrew~Gordon Wilson, and Micah Goldblum.
\newblock Transfer learning with deep tabular models.
\newblock \emph{arXiv preprint arXiv:2206.15306}, 2022.

\bibitem[Liang et~al.(2022)Liang, Wang, Gao, Wang, Zhao, and
  Wang]{liang2022self}
Dong Liang, Jun Wang, Xiaoyu Gao, Jiahui Wang, Xiaoyong Zhao, and Lei Wang.
\newblock Self-supervised pretraining isolated forest for outlier detection.
\newblock In \emph{2022 International Conference on Big Data, Information and
  Computer Network (BDICN)}, pp.\  306--310. IEEE, 2022.

\bibitem[Liu \& Avci(2019)Liu and Avci]{liu2019incorporating}
Frederick Liu and Besim Avci.
\newblock Incorporating priors with feature attribution on text classification.
\newblock \emph{arXiv preprint arXiv:1906.08286}, 2019.

\bibitem[Liu et~al.(2021)Liu, Lin, Liu, Rehg, Paull, Xiong, Song, and
  Weller]{liu2021orthogonal}
Weiyang Liu, Rongmei Lin, Zhen Liu, James~M Rehg, Liam Paull, Li~Xiong,
  Le~Song, and Adrian Weller.
\newblock Orthogonal over-parameterized training.
\newblock In \emph{Proceedings of the IEEE/CVF Conference on Computer Vision
  and Pattern Recognition}, pp.\  7251--7260, 2021.

\bibitem[Loshchilov \& Hutter(2017)Loshchilov and
  Hutter]{loshchilov2017decoupled}
Ilya Loshchilov and Frank Hutter.
\newblock Decoupled weight decay regularization.
\newblock \emph{arXiv preprint arXiv:1711.05101}, 2017.

\bibitem[Lundberg \& Lee(2017)Lundberg and Lee]{lundberg2017unified}
Scott~M Lundberg and Su-In Lee.
\newblock A unified approach to interpreting model predictions.
\newblock \emph{Advances in Neural Information Processing Systems}, 30, 2017.

\bibitem[Luo et~al.(2016)Luo, Li, Urtasun, and Zemel]{luo2016understanding}
Wenjie Luo, Yujia Li, Raquel Urtasun, and Richard Zemel.
\newblock Understanding the effective receptive field in deep convolutional
  neural networks.
\newblock \emph{Advances in Neural Information Processing Systems}, 29, 2016.

\bibitem[Michalski et~al.(1986)Michalski, Mozetic, Hong, and
  Lavrac]{michalski1986multi}
Ryszard~S Michalski, Igor Mozetic, Jiarong Hong, and Nada Lavrac.
\newblock The multi-purpose incremental learning system aq15 and its testing
  application to three medical domains.
\newblock In \emph{Proc. AAAI}, volume 1986, pp.\  1--041, 1986.

\bibitem[Moosavi-Dezfooli et~al.(2019)Moosavi-Dezfooli, Fawzi, Uesato, and
  Frossard]{moosavi2019robustness}
Seyed-Mohsen Moosavi-Dezfooli, Alhussein Fawzi, Jonathan Uesato, and Pascal
  Frossard.
\newblock Robustness via curvature regularization, and vice versa.
\newblock In \emph{Proceedings of the IEEE/CVF Conference on Computer Vision
  and Pattern Recognition}, pp.\  9078--9086, 2019.

\bibitem[Moro et~al.(2016)Moro, Rita, and Vala]{moro2016predicting}
S{\'e}rgio Moro, Paulo Rita, and Bernardo Vala.
\newblock Predicting social media performance metrics and evaluation of the
  impact on brand building: A data mining approach.
\newblock \emph{Journal of Business Research}, 69\penalty0 (9):\penalty0
  3341--3351, 2016.

\bibitem[Paszke et~al.(2019)Paszke, Gross, Massa, Lerer, Bradbury, Chanan,
  Killeen, Lin, Gimelshein, Antiga, et~al.]{paszke2019pytorch}
Adam Paszke, Sam Gross, Francisco Massa, Adam Lerer, James Bradbury, Gregory
  Chanan, Trevor Killeen, Zeming Lin, Natalia Gimelshein, Luca Antiga, et~al.
\newblock {PyTorch}: An imperative style, high-performance deep learning
  library.
\newblock \emph{Advances in Neural Information Processing Systems}, 32, 2019.

\bibitem[Pereyra et~al.(2017)Pereyra, Tucker, Chorowski, Kaiser, and
  Hinton]{pereyra2017regularizing}
Gabriel Pereyra, George Tucker, Jan Chorowski, {\L}ukasz Kaiser, and Geoffrey
  Hinton.
\newblock Regularizing neural networks by penalizing confident output
  distributions.
\newblock \emph{arXiv preprint arXiv:1701.06548}, 2017.

\bibitem[Prokhorenkova et~al.(2018)Prokhorenkova, Gusev, Vorobev, Dorogush, and
  Gulin]{prokhorenkova2018catboost}
Liudmila Prokhorenkova, Gleb Gusev, Aleksandr Vorobev, Anna~Veronika Dorogush,
  and Andrey Gulin.
\newblock Catboost: unbiased boosting with categorical features.
\newblock \emph{Advances in Neural Information Processing Systems}, 31, 2018.

\bibitem[Quinlan(1987)]{quinlan1987simplifying}
J.~Ross Quinlan.
\newblock Simplifying decision trees.
\newblock \emph{International journal of man-machine studies}, 27\penalty0
  (3):\penalty0 221--234, 1987.

\bibitem[Ramachandram \& Taylor(2017)Ramachandram and
  Taylor]{ramachandram2017deep}
Dhanesh Ramachandram and Graham~W Taylor.
\newblock Deep multimodal learning: A survey on recent advances and trends.
\newblock \emph{IEEE Signal Processing Magazine}, 34\penalty0 (6):\penalty0
  96--108, 2017.

\bibitem[Rifai et~al.(2011)Rifai, Vincent, Muller, Glorot, and
  Bengio]{rifai2011contractive}
Salah Rifai, Pascal Vincent, Xavier Muller, Xavier Glorot, and Yoshua Bengio.
\newblock Contractive auto-encoders: Explicit invariance during feature
  extraction.
\newblock In \emph{Icml}, 2011.

\bibitem[Ross et~al.(2017{\natexlab{a}})Ross, Lage, and
  Doshi-Velez]{ross2017neural}
Andrew Ross, Isaac Lage, and Finale Doshi-Velez.
\newblock The neural {LASSO}: Local linear sparsity for interpretable
  explanations.
\newblock In \emph{Workshop on Transparent and Interpretable Machine Learning
  in Safety Critical Environments, 31st Conference on Neural Information
  Processing Systems}, volume~4, 2017{\natexlab{a}}.

\bibitem[Ross et~al.(2017{\natexlab{b}})Ross, Hughes, and
  Doshi-Velez]{ross2017right}
Andrew~Slavin Ross, Michael~C Hughes, and Finale Doshi-Velez.
\newblock Right for the right reasons: Training differentiable models by
  constraining their explanations.
\newblock \emph{arXiv preprint arXiv:1703.03717}, 2017{\natexlab{b}}.

\bibitem[Rubachev et~al.(2022)Rubachev, Alekberov, Gorishniy, and
  Babenko]{rubachev2022revisiting}
Ivan Rubachev, Artem Alekberov, Yury Gorishniy, and Artem Babenko.
\newblock Revisiting pretraining objectives for tabular deep learning.
\newblock \emph{arXiv preprint arXiv:2207.03208}, 2022.

\bibitem[Seedat et~al.(2023)Seedat, Jeffares, Imrie, and van~der
  Schaar]{seedat2023improving}
Nabeel Seedat, Alan Jeffares, Fergus Imrie, and Mihaela van~der Schaar.
\newblock Improving adaptive conformal prediction using self-supervised
  learning.
\newblock \emph{arXiv preprint arXiv:2302.12238}, 2023.

\bibitem[Shwartz-Ziv \& Armon(2022)Shwartz-Ziv and Armon]{shwartz2022tabular}
Ravid Shwartz-Ziv and Amitai Armon.
\newblock Tabular data: Deep learning is not all you need.
\newblock \emph{Information Fusion}, 81:\penalty0 84--90, 2022.

\bibitem[Street et~al.(1993)Street, Wolberg, and
  Mangasarian]{street1993nuclear}
W~Nick Street, William~H Wolberg, and Olvi~L Mangasarian.
\newblock Nuclear feature extraction for breast tumor diagnosis.
\newblock In \emph{Biomedical Image Processing and Biomedical Visualization},
  volume 1905, pp.\  861--870. SPIE, 1993.

\bibitem[Sun et~al.(2019)Sun, Yang, Zhang, Lin, Dong, Young, and
  Dong]{sun2019supertml}
Baohua Sun, Lin Yang, Wenhan Zhang, Michael Lin, Patrick Dong, Charles Young,
  and Jason Dong.
\newblock {SuperTML}: Two-dimensional word embedding for the precognition on
  structured tabular data.
\newblock In \emph{Proceedings of the IEEE/CVF Conference on Computer Vision
  and Pattern Recognition Workshops}, pp.\  0--0, 2019.

\bibitem[Sundararajan et~al.(2017)Sundararajan, Taly, and
  Yan]{sundararajan2017axiomatic}
Mukund Sundararajan, Ankur Taly, and Qiqi Yan.
\newblock Axiomatic attribution for deep networks.
\newblock In \emph{International conference on machine learning}, pp.\
  3319--3328. PMLR, 2017.

\bibitem[Tang et~al.(2020)Tang, Kumar, Chen, and Shrivastava]{tang2020deep}
Michelle Tang, Pulkit Kumar, Hao Chen, and Abhinav Shrivastava.
\newblock Deep multimodal learning for the diagnosis of autism spectrum
  disorder.
\newblock \emph{Journal of Imaging}, 6\penalty0 (6):\penalty0 47, 2020.

\bibitem[Thompson et~al.(2013)Thompson, Blair, Chen, and
  Henrey]{thompson2013video}
Joseph~J Thompson, Mark~R Blair, Lihan Chen, and Andrew~J Henrey.
\newblock Video game telemetry as a critical tool in the study of complex skill
  learning.
\newblock \emph{PloS one}, 8\penalty0 (9):\penalty0 e75129, 2013.

\bibitem[Tibshirani(1996)]{tibshirani1996regression}
Robert Tibshirani.
\newblock Regression shrinkage and selection via the lasso.
\newblock \emph{Journal of the Royal Statistical Society: Series B
  (Methodological)}, 58\penalty0 (1):\penalty0 267--288, 1996.

\bibitem[Ucar et~al.(2021)Ucar, Hajiramezanali, and Edwards]{ucar2021subtab}
Talip Ucar, Ehsan Hajiramezanali, and Lindsay Edwards.
\newblock {SubTab}: Subsetting features of tabular data for self-supervised
  representation learning.
\newblock \emph{Advances in Neural Information Processing Systems},
  34:\penalty0 18853--18865, 2021.

\bibitem[Vaswani et~al.(2017)Vaswani, Shazeer, Parmar, Uszkoreit, Jones, Gomez,
  Kaiser, and Polosukhin]{vaswani2017attention}
Ashish Vaswani, Noam Shazeer, Niki Parmar, Jakob Uszkoreit, Llion Jones,
  Aidan~N Gomez, {\L}ukasz Kaiser, and Illia Polosukhin.
\newblock Attention is all you need.
\newblock \emph{Advances in Neural Information Processing Systems}, 30, 2017.

\bibitem[Wang \& Sun(2022)Wang and Sun]{wang2022transtab}
Zifeng Wang and Jimeng Sun.
\newblock {TransTab}: Learning transferable tabular transformers across tables.
\newblock \emph{Advances in Neural Information Processing Systems}, 2022.
\newblock URL \url{https://openreview.net/forum?id=A1yGs_SWiIi}.

\bibitem[Waugh(1995)]{waugh1995extending}
Samuel~George Waugh.
\newblock \emph{Extending and benchmarking Cascade-Correlation: extensions to
  the Cascade-Correlation architecture and benchmarking of feed-forward
  supervised artificial neural networks}.
\newblock PhD thesis, University of Tasmania, 1995.

\bibitem[Wen et~al.(2020)Wen, Tran, and Ba]{wen2020batchensemble}
Yeming Wen, Dustin Tran, and Jimmy Ba.
\newblock {BatchEnsemble}: an alternative approach to efficient ensemble and
  lifelong learning.
\newblock \emph{arXiv preprint arXiv:2002.06715}, 2020.

\bibitem[Wilcoxon(1992)]{wilcoxon1992individual}
Frank Wilcoxon.
\newblock Individual comparisons by ranking methods.
\newblock In \emph{Breakthroughs in Statistics}, pp.\  196--202. Springer,
  1992.

\bibitem[Wu et~al.(2022)Wu, Li, Cui, and Xu]{wu2022deep}
Xinglong Wu, Mengying Li, Xin-wu Cui, and Guoping Xu.
\newblock Deep multimodal learning for lymph node metastasis prediction of
  primary thyroid cancer.
\newblock \emph{Physics in Medicine \& Biology}, 67\penalty0 (3):\penalty0
  035008, 2022.

\bibitem[Yoon et~al.(2020)Yoon, Zhang, Jordon, and van~der
  Schaar]{yoon2020vime}
Jinsung Yoon, Yao Zhang, James Jordon, and Mihaela van~der Schaar.
\newblock {VIME}: Extending the success of self-and semi-supervised learning to
  tabular domain.
\newblock \emph{Advances in Neural Information Processing Systems},
  33:\penalty0 11033--11043, 2020.

\bibitem[Zhang et~al.(2018)Zhang, Cisse, Dauphin, and
  Lopez-Paz]{zhang2017mixup}
Hongyi Zhang, Moustapha Cisse, Yann~N Dauphin, and David Lopez-Paz.
\newblock mixup: Beyond empirical risk minimization.
\newblock In \emph{International Conference on Learning Representations}, 2018.

\bibitem[Zhang et~al.(2021)Zhang, Ti{\v{n}}o, Leonardis, and
  Tang]{zhang2021survey}
Yu~Zhang, Peter Ti{\v{n}}o, Ale{\v{s}} Leonardis, and Ke~Tang.
\newblock A survey on neural network interpretability.
\newblock \emph{IEEE Transactions on Emerging Topics in Computational
  Intelligence}, 2021.

\bibitem[Zhu et~al.(2021)Zhu, Brettin, Xia, Partin, Shukla, Yoo, Evrard,
  Doroshow, and Stevens]{zhu2021converting}
Yitan Zhu, Thomas Brettin, Fangfang Xia, Alexander Partin, Maulik Shukla,
  Hyunseung Yoo, Yvonne~A Evrard, James~H Doroshow, and Rick~L Stevens.
\newblock Converting tabular data into images for deep learning with
  convolutional neural networks.
\newblock \emph{Scientific reports}, 11\penalty0 (1):\penalty0 11325, 2021.

\bibitem[Zi{k{e}}ba et~al.(2013)Zi{k{e}}ba, Tomczak, Lubicz, and
  {'S}wi{k{a}}tek]{zieba2013boosted}
Maciej Zi{k{e}}ba, Jakub~M Tomczak, Marek Lubicz, and Jerzy {'S}wi{k{a}}tek.
\newblock Boosted {SVM} for extracting rules from imbalanced data in
  application to prediction of the post-operative life expectancy in the lung
  cancer patients.
\newblock \emph{Applied Soft Computing}, 2013.

\bibitem[Zou \& Hastie(2005)Zou and Hastie]{zou2005regularization}
Hui Zou and Trevor Hastie.
\newblock Regularization and variable selection via the elastic net.
\newblock \emph{Journal of the royal statistical society: series B (statistical
  methodology)}, 67\penalty0 (2):\penalty0 301--320, 2005.

\end{thebibliography}
